\documentclass[UTF8]{article} 
\usepackage{iclr2023_conference,times}
\pdfoutput=1 

\usepackage{amsmath,amsfonts,bm}

\def\eqref#1{equation~\ref{#1}}

\def\1{\bm{1}}

\DeclareMathAlphabet{\mathsfit}{\encodingdefault}{\sfdefault}{m}{sl}
\SetMathAlphabet{\mathsfit}{bold}{\encodingdefault}{\sfdefault}{bx}{n}

\usepackage{microtype}
\usepackage{amsmath}
\usepackage{booktabs}
\usepackage{colortbl}
\usepackage{CJKutf8}
\usepackage[vietnamese, main=english]{babel}
\usepackage[T1, T5]{fontenc}
\usepackage[utf8]{inputenc}
\definecolor{lightgray}{rgb}{0.9,0.9,0.9}
\usepackage{caption}
\usepackage{subcaption}
\usepackage{xcolor}
\usepackage{graphicx}
\usepackage{setspace}
\usepackage{hyperref}
\usepackage{url}
\usepackage{multirow}
\usepackage{colortbl}
\usepackage{tabularx}
\usepackage{blindtext}
\usepackage{pgfplots}
\pgfplotsset{compat=1.18} 
\usepackage{tikz}
\usetikzlibrary{er,positioning,bayesnet}
\usepackage[inline]{enumitem}
\usepackage{makecell}
\usepackage{tipa}
\usepackage{siunitx}
\usepackage{svg}
\usepackage{tocloft}
\usepackage{listings}
\usepackage[raster,skins,most,breakable]{tcolorbox} 
\usepackage{xltabular}
\usepackage[framemethod=tikz]{mdframed}
\usepackage{diagbox}
\usepackage{graphicx}
\usepackage{listingsutf8}
\usepackage{verbatim}
\surroundwithmdframed[
  hidealllines=true,
  innerleftmargin=0pt,
  innertopmargin=0pt,
  innerbottommargin=0pt]{lstlisting}
\lstnewenvironment{response}[1][] 
 {\lstset{
 columns=fullflexible,
  breakautoindent=false, breakindent=0pt, breaklines, linewidth=8cm, #1}}
 {}
\tcbuselibrary{breakable}

\title{Compass-V2 Technical Report}
\author{ \\
\centerline {\textbf {Sophia Maria}}
\\
\and
\centerline {{\tt sophia.maria@shopee.com } }
\\ 
\and
\centerline {Shopee LLM Team}
\and
\centerline {2025-04} \\
}
\usepackage{pifont}

\iclrfinalcopy
\begin{document}

\maketitle


\begin{abstract}
Predominant LLMs focus on high-resource languages while leaving low-resource languages, particularly those in Southeast Asia (SEA), underrepresented. In addition, those models are general-purpose and pay limited attention to the e-commerce domain. To overcome these limitations, we introduce Compass-v2, a lightweight Mixture-of-Experts (MoE) model specifically designed for Southeast Asian languages and e-commerce applications. To balance model performance and inference cost, the model is designed with 30B total parameters and 5B active parameters, incorporating both fine-grained and shared expert modules. To enhance multilingual performance, we curated and constructed a high-quality, industry-leading SEA dataset, to the best of our knowledge. To boost performance in the e-commerce domain, we built a dataset comprising hundreds of billions of tokens, sourced through external data mining and internal platform collection. Besides, we pioneered a hybrid reasoning model that supports both fast thinking and deep thinking within a unified framework to enhance the reasoning capabilities, diverging from the conventional industry practice of deploying two separate models. Through extensive experimental evaluations, our model demonstrates state-of-the-art SEA multilingual and e-commerce performance among sub-30B models, while maintaining significantly lower inference cost.

\begin{figure}[h]
    \centering
    \begin{minipage}{0.6\textwidth}
        \centering
        \includegraphics[width=\linewidth]{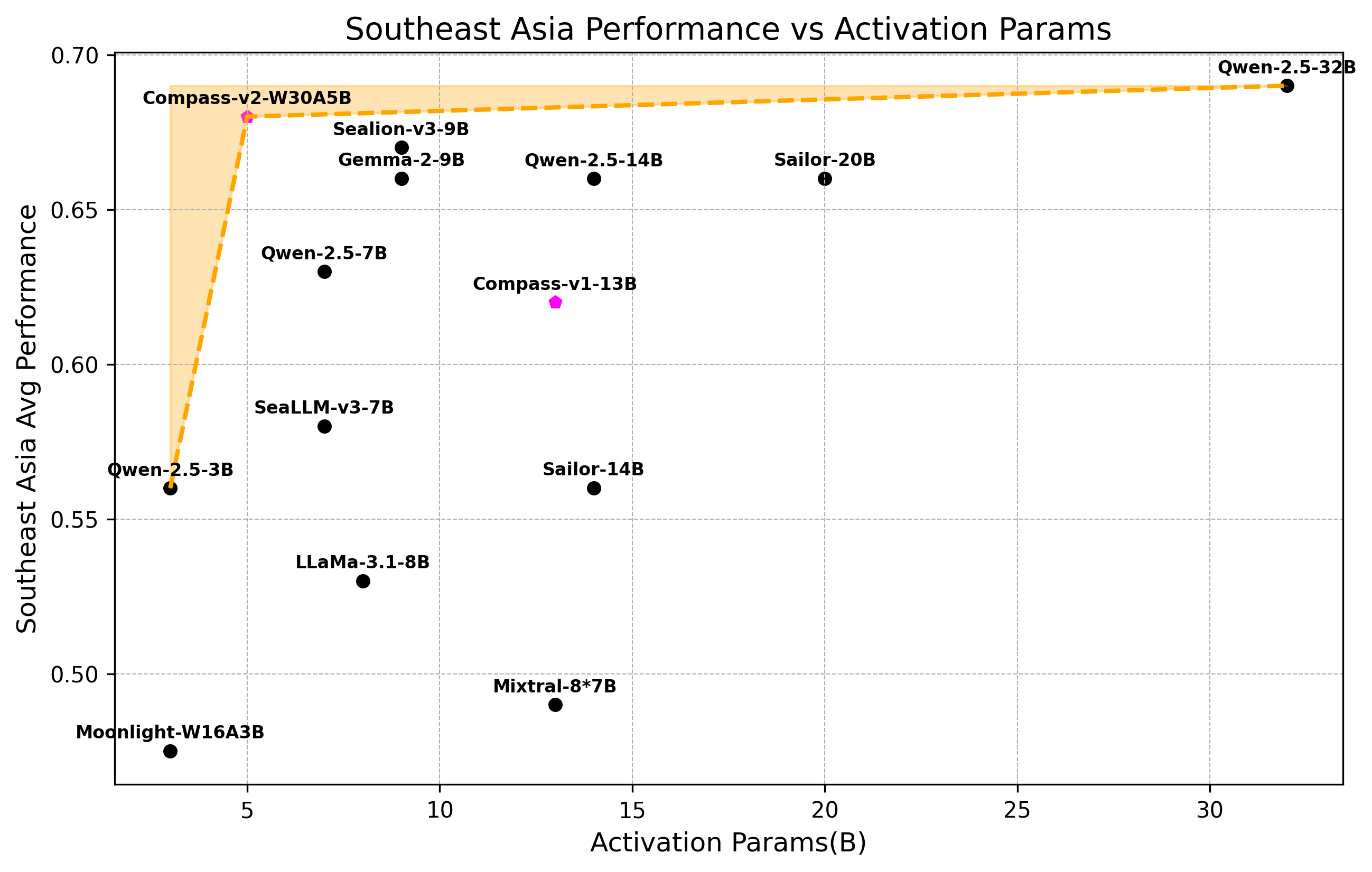}
        \caption{Southeast Asia Performance comparison}
    \end{minipage}
    \begin{minipage}{0.35\textwidth}
        \centering
        \includegraphics[width=\linewidth]{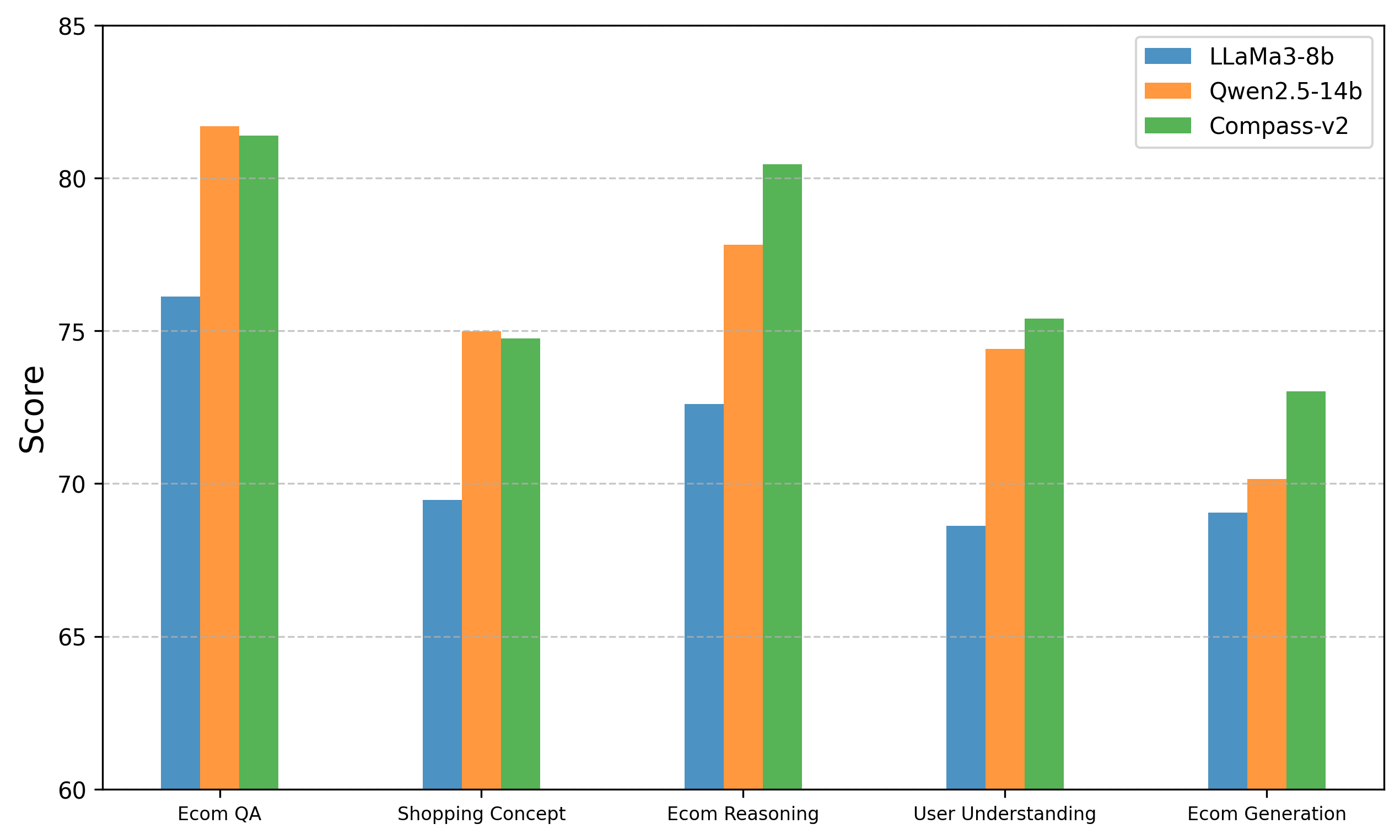}
        \caption{Ecommerce comparison of LLaMa3-8b, Qwen2.5-14b, and Compass-v2}
    \end{minipage}
\end{figure}

\vfill



\end{abstract}
\clearpage

\tableofcontents
\clearpage

\section{Introduction}

Large Language Models (LLMs) have seen significant advancements in recent years, becoming foundational technologies across a wide range of natural language process (NLP) applications such as dialogue systems, code generation, and document understanding. Breakthrough models including GPT series~\citep{Achiam2023GPT4TR}, LLaMA series~\citep{Dubey2024TheL3}, Qwen series~\citep{Yang2024Qwen25TR} and DeepSeek series~\citep{DeepSeekAI2024DeepSeekV3TR} have pushed the boundaries of LLM by scaling up model size and training data. 

However, these models have largely focused on high-resource language such as English and Chinese, leaving low-resource languages (particular those spoken in Southeast Asia) underrepresented. This presents a growing challenge especially in real-world applications like e-commerce, where understanding multilingual user queries and culturally specific expressions is crucial. Additionally, many LLMs are designed either for lightweight, fast-thinking responses or for slow, deep-thinking tasks, but few integrate both capabilities in a unified manner.

To address these limitations, we introduce \textbf{Compass-v2}, a hybrid reasoning Mixture-of-Experts (MoE) language model tailored for Southeast Asian languages and e-commerce scenarios. Unlike many existing LLMs that rely on continued pre-training from general-purpose checkpoints~\citep{sailor2report,Nguyen2023SeaLLMsL}, our model is trained from scratch using curated multilingual and domain-specific datasets such as e-commerce data. The train-from-scratch approach allows us to precisely control the data distribution, tokenization design and model capacity to better capture linguistic diversity and commercial nuances of Southeast Asia. Notably, the Compass-v2 tokenizer has been optimized to
achieve the best compression rate for SEA languages, enabling more efficient encoding of multilingual inputs.

Compass-v2 strikes an effective balance between fast thinking and deep thinking: it supports shallow reasoning with low latency for lightweight interactions, while also enabling complex, long-context reasoning when deeper understanding is required. This is particularly valuable in multilingual and domain-specific settings, such as Southeast Asian e-commerce, where real-time performance and contextual accuracy must coexist. To achieve this, Compass-v2 incorporates advancements from both the data and architecture perspectives.

From the data perspective, Compass-v2 combines high-quality, diverse datasets—including large-scale multilingual corpora and e-commerce-specific data—with long chain-of-thought (CoT) samples that contain both the step-by-step reasoning process and corresponding summarized answers. The former enhances the model’s ability to generalize across languages and domains, particularly in multilingual and commercial contexts, while the latter strengthens its capacity for deep, multi-step reasoning, enabling more accurate and coherent responses in complex tasks.


From the architecture perspective, Compass-v2 adopts a fine-grained Mixture-of-Experts (MoE) architecture with shared experts~\citep{Shazeer2017OutrageouslyLN}. 
This design enables a larger number of expert combinations without increasing computational cost, improving both expert utilization and expressiveness~\citep{dai2024deepseekmoe, Yang2024Qwen25TR}.  In Compass-v2, each token is routed to 2 shared experts and 4 specialized experts. This hybrid expert assignment supports both generalization and specialization, enabling the model to handle a wide range of tasks efficiently.

During pre-training, we first train Compass-v2 on 8T diverse tokens from a diverse corpus. To improve training stability and refine the model's capabilities, we then perform an annealing stage on 4T high-quality tokens to stabilize the training process. Next, we conduct a context length extension stage, increasing the maximum context window from 4k to 32k, which equips the model with long-context reasoning abilities. After pre-training, we carry out two supervised fine-tuning (SFT) stages on 6 million instruction-following samples, and then we apply direct preference optimization (DPO) to align the Compass-v2 base model with human preferences and further unlock its potential.

We conduct a comprehensive evaluations on Compass-v2 across a range of capabilities. In addition to standard open-source benchmarks, we also design a series of in-house evaluation datasets derived from real-world scenarios to assess the performance of Compass-v2 in multilingual, e-commerce, business and general-purpose capabilities. 
The evaluation results show that, despite its significantly smaller parameter size, Compass-v2 outperforms models of similar scale in multilingual, e-commerce, and business-related tasks. In open-source e-commerce and General Field evaluation, Comapss-v2 beats the second best competing model by 10.6\% and 0.5\% respectively. Notably, Compass-v2 achieves comparable performance to the industry’s top models, such as GPT-4o and Qwem2.5-72b, on the in-house dataset despite being much smaller in scale.  This highlights its strong efficiency-to-performance trade-off, making it a competitive choice for practical deployment under constrained resources. Within Shopee's ecosystem, Compass-v2 powers a variety of downstream applications including search and recommendation, video chat, and live-streaming support, driving improvements in user intent understanding, conversion rates, and customer engagement.

To reduce the GPU memory demands during inference, we explore low-precision quantization techniques, including FP8 and Activation-aware Weight Quantizaiton (AWQ). Our evaluations show that while FP8 provides moderate improves under certain conditions, AWQ achieves consistently superior efficiency, particularly for large-batch inference. Moreover, both quantization approaches maintain performance comparable to FP16 across multiple key benchmarks. To facilitate easy access and integration, we deploy Compass-v2 through CAP -- Shopee's internal MaaS platform. We offers two access methods for the Compass series of models: 1) a web portal for rapid experimentation and testing, and 2) an API for integrating Compass models into real-world business workflows. This architecture enables enterprises to efficiently adopt and scale advanced LLM capabilities without the overhead of model development, making Compass-v2 not only efficient in inference but also practical in real-world deployment scenarios.

Our contributions are summarized as follows:
\begin{itemize}
    \item We design Compass-v2, a novel hybrid reasoning MoE model with 30B total parameters and 5B active parameters, incorporating fine-grained and shared expert modules. To enhance multilingual and e-commerce performance, we curated an industry-leading SEA dataset and constructed an e-commerce corpus of hundreds of billions of tokens from diverse sources.

    \item We pioneered a hybrid reasoning model that supports both fast thinking and deep thinking within a unified framework to enhance the reasoning capabilities.

    \item Compass-v2 outperforms models of similar scale in multilingual, e-commerce, and business-related tasks. In open-source e-commerce and General Field evaluation, Comapss-v2 beats the second best model by 10.6\% and 0.5\% respectively. Notably, Compass-v2 achieves comparable performance to the industry’s top models, such as GPT-4o and Qwem2.5-72b, on the in-house dataset despite being much smaller in scale. 

    \item We release the Compass API for internal use within the company.
\end{itemize}

The rest of this paper is structured as follows:
\begin{itemize}
    \item In Section~\ref{chapeter:pretraining}, we outline our pre-training stages, which comprises a primary pre-training stage followed by two continuous pre-training stages: an annealing stage and a long-context training stage. We describe our extensive data collection and curation process, coupled with a tokenizer specifically optimized for the unique needs of a small MoE model.
    \item In Section~\ref{chapter:posttraining} and Section~\ref{chapter:DPO}, we describe our post-training process, which includes two supervised fine-tuning stages and one direct preference optimization (DPO) stage. We elaborate on the details of the supervised training approach and explain how we construct our fine-tuning dataset.
    \item In Section~\ref{chapter:inference}, we present our inference acceleration and model quantization techniques for the Compass-v2 model, focusing on addressing latency and memory bottlenecks.
    \item In Section~\ref{chapter:evaluation}, we discuss the benchmarks, baselines, and evaluation pipelines employed to assess the performance of our approach. We perform rigorous testing against multiple benchmarks and baselines to demonstrate the effectiveness of the proposed small MoE LLM in terms of accuracy, efficiency, and scalability.
\end{itemize}

\section{Pre-Training} \label{chapeter:pretraining}

\subsection{Pretraining Data}\label{sub:pretraining:data}
The pretraining data in Compass-v2 model encompasses a wide diversity of sources in terms of the languages and document types covered, resulting in a high-quality and diverse corpus comprising 12 trillion tokens. On top of the mainstream languages like English and Chinese, we have particularly focused our efforts on enlarging our data corpus for Southeast Asian languages, comprising of Indonesian, Thai, Vietnamese, Malay, Tagalog, as well as Portuguese to support our businesses in Southeast Asia and South America. Furthermore, the Compass-v2 tokenizer has been optimized to achieve the best compression rate for SEA languages, enabling more efficient encoding of multilingual inputs and better utilization of model capacity.


\subsubsection{Industry-leading SEA data}
\begin{figure}[htbp]
    \centering
    \begin{minipage}{0.89\textwidth} 
        \centering
        \includegraphics[width=\textwidth]{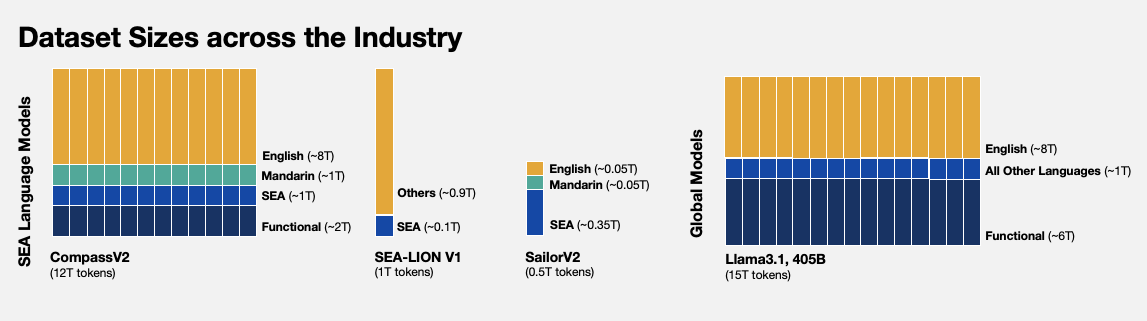}
        \caption{Industry Overview of Pretraining Dataset Scale.}
        \label{fig:IndustryOverview}
    \end{minipage}    
\end{figure}

To enhance the model’s capabilities in Southeast Asian (SEA) languages, Portuguese, and the e-commerce domain, we significantly expanded our pretraining corpus by incorporating a wide range of relevant and high-quality data sources. Among these, SEA data accounts for a significant portion of the Compass-v2 corpus. As shown in Figure~\ref{fig:IndustryOverview}, Compass-v2 surpasses most existing open-source SEA language model pretraining corpora in data richness.

\begin{figure}[htbp]
    \centering

    \begin{minipage}{0.43\textwidth} 
        \centering
        \includegraphics[width=1.1\textwidth]{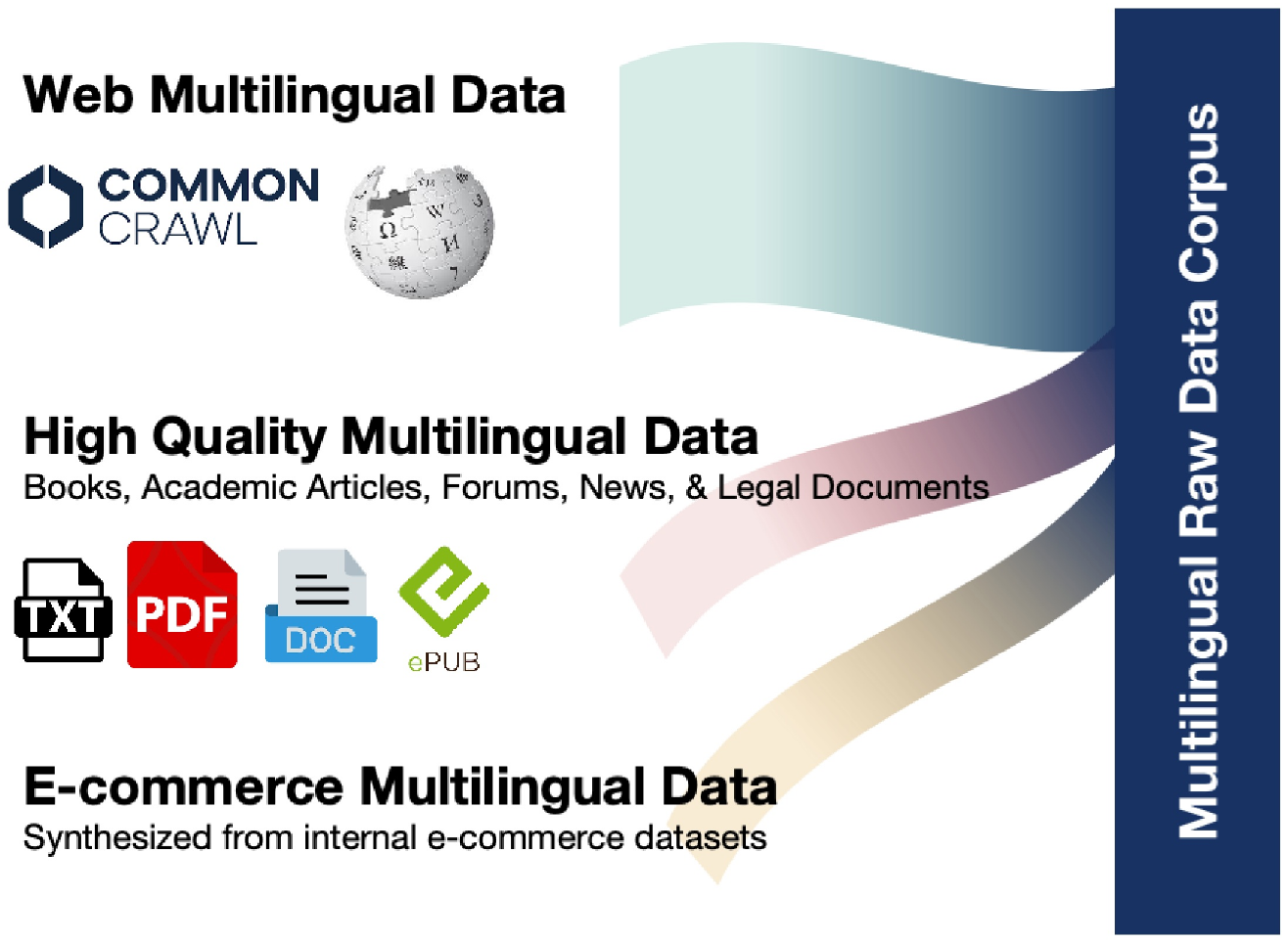}
        \caption{Multilingual Corpus Construction}
        \label{fig:MultilingualCorpus}
    \end{minipage}
    \begin{minipage}{0.56\textwidth} 
        \centering
        \includegraphics[width=\textwidth]{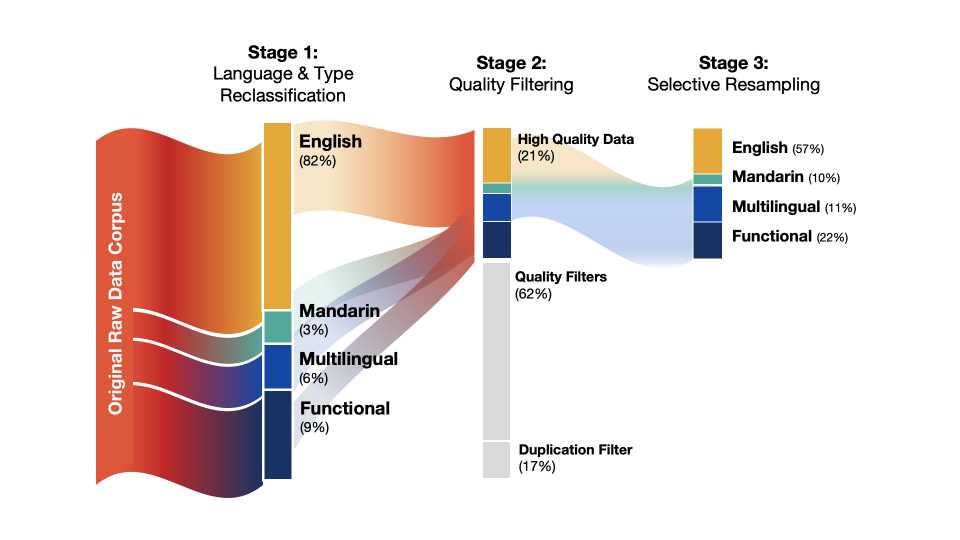}
        \caption{Pretraining Data Processing Overview.}
        \label{fig:DataProcessingOverview}
    \end{minipage}
    
\end{figure}

\textbf{Industry-leading SEA datasets} Our SEA data construction follows a comprehensive, multi-stage pipeline for parsing and decoding Common Crawl, Wikipedia, high quality documents, and synthesizing relevant e-commerce documents, as shown in Figure ~\ref{fig:MultilingualCorpus}. We first parsed the full Common Crawl dataset \citep{commoncrawl} to extract vast and diverse set of target language data. To further enrich the corpus with high-quality document-style content, we extracted text from PDF, TXT, DOC, and EPUB files, focusing on topics related to Southeast Asian culture, history, and education. In addition, we obtained high-quality content from external sources, including corpora in Thai, Indonesian, Vietnamese, and Malay. These datasets span a variety of formats such as multi-turn dialogues, news articles, books, and academic publications, enhancing the domain diversity of the training data.

\textbf{Massive e-commerce datasets} To address domain-specific needs, particularly in e-commerce, we integrated a substantial volume of internal Shopee data, encompassing a wide range of user interaction scenarios such as product search, reviews, chats, and descriptions. In addition to real-world platform data, we also synthesized domain-specific datasets including multilingual product search queries, descriptions, and user-generated reviews. This dual approach enables us to construct a dataset that enhances user intent capture and improves recommendation quality in SEA markets.

Through this carefully designed data acquisition pipeline, Compass-v2 offers a large-scale, diverse, and high-quality dataset optimized for SEA market and practical business applications.

\subsubsection{State-of-the-art data quality}


Building upon the pretraining data curation pipeline established in our first-generation Compass model~\citep{Maria2024CompassLM}, we further enhance the quality of our dataset by focusing on how to customize our curation process for low-resource languages, code, math, and e-commerce domains. The new Compass-v2 dataset and processing pipeline as illustrated in Figure~\ref{fig:DataProcessingOverview}. The pipeline results in substantial improvements in pretraining data quality over its predecessor~\citep{Maria2024CompassLM}, driven by several key factors: 

\textbf{Deduplication} While deduplication is often considered a standard best practice, our experimental results in Table~\ref{tab:comparison_of_en_dataset} show that excessively deduplicated versions of DCLM (EN) and Combined (EN) perform worse on average compared to their less deduplicated versions. This suggests that repeated high-quality examples may play a positive role in reinforcing model behavior, particularly for developing emergent capabilities such as complex reasoning and instruction-following. Hence, we carefully retained high-quality repetition to help models better internalize consistent patterns, especially in limited-resource or domain-specific contexts.

\textbf{Heuristic filtering} Recognizing the importance of code and math-related content, we apply a separate set of heuristics to filter these domains independently. This ensures that the texts with a higher similarity to Code and Math is not disproportionately filtered out during the general quality selection process. 

\textbf{Model-based filtering} To improve overall data quality across all languages, we employ a variety of small models and classifiers~\citep{NEURIPS2024_19e4ea30, NEURIPS2024_370df50c, Su2024NemotronCCTC, Conneau2019UnsupervisedCR} to identify texts with high fluency, educational value, and linguistic coherence. These classifiers help us retain samples that contribute to better generalization and instruction-following capabilities in the final model. 

\textbf{Selective resampling} To further balance diversity and quality, we perform a selective resampling strategy that combines quality scores with the original occurrence distributions of documents. This allows us to preserve the natural variability of the raw corpus while promoting higher-quality samples in the final pretraining dataset.



\begin{table}[htbp]
\resizebox*{1\linewidth}{!}{
\begin{tabular}{lcccccccc|c}
\toprule
\textbf{Dataset}                    & \textbf{MMLU}   & \textbf{BoolQ}  & \textbf{ARC\_E} & \textbf{ARC\_C} & \textbf{HelloSwag} & \textbf{OpenBookQA} & \textbf{PIQA}   & \textbf{Winogrande} & \textbf{Average} \\ 
\midrule
DCLM (EN)                   & 0.4811 & 0.7171 & 0.7403 & 0.4505 & 0.7413    & 0.420      & 0.7894 & 0.6740     & 0.6291           \\
DCLM (EN), more deduplication     & 0.4523 & 0.7162 & 0.7340 & 0.4548 & 0.7328    & 0.446      & 0.7862 & 0.6701     & 0.6264           \\
Combined (EN)               & 0.4955 & 0.7410 & 0.7475 & 0.4710 & 0.7400    & 0.410      & 0.7976 & 0.6709     & 0.6366           \\
Combined (EN), more deduplication & 0.4454 & 0.7107 & 0.7340 & 0.4548 & 0.7257    & 0.414      & 0.7873 & 0.6488     & 0.6175           \\
\midrule
Compass-v2 (EN)          & 0.5032 & 0.7372 & 0.7412 & 0.4812 & 0.7408    & 0.426      & 0.7943 & 0.6740     & \textbf{0.6397}  \\
\bottomrule
\end{tabular}
}
\caption{Performance comparison among Compass-v2 (EN), DCLM (EN), Combined (EN) on core English-related benchmarks. Combined (EN) is an English mixture of DCLM, FineWeb-Edu and Nemotron-CC.}
\label{tab:comparison_of_en_dataset}
\end{table}

With these enhancements, we have a larger and high-quality pre-training corpus that surpasses the most advanced open-source datasets in the industry.

\textbf{Compass-v2 dataset surpasses the best open-source datasets.} To better validate the quality of our final dataset, we conducted several online evaluations to compare Compass-v2 English pretraining data against several state-of-the-art open-source, high-quality datasets, such as DCLM~\citep{Li2024DataCompLMIS}, FineWeb-Edu~\citep{NEURIPS2024_370df50c}, and Nemotron-CC~\citep{Su2024NemotronCCTC}.
Following the setting in ~\cite{Li2024DataCompLMIS}, we sampled 210B tokens from each dataset and trained a 7B-parameter model under identical training settings. As shown in Table~\ref{tab:comparison_of_en_dataset}, Compass-v2 (EN) dataset achieves the highest average score of 0.6397, outperforming both the DCLM (EN) dataset and the Combined (EN)  mixture that integrates high-quality samples from DCLM, FineWeb-Edu, and Nemotron-CC. This results clearly demonstrate that the Compass-v2 English dataset surpasses the best open-source datasets in the industry. The results highlight the strength of our data curation pipeline, particularly in preserving high-quality samples and maintaining domain and linguistic diversity.



\subsubsection{SEA-adapted tokenization}

We focus on the performance of the model across multiple languages, including mainstream languages, Southeast Asian languages, and some low-resource languages.
To achieve this, the tokenizer must maintain a good compression rate in different languages while keeping the vocabulary size appropriate to avoid excessive embedding parameters.
We carefully design the tokenizer to strike this balance. 

The tokenizer employs the byte pair encoding (BPE) algorithm \citep{sennrich-etal-2016-neural} and is trained through SentencePiece \citep{kudo-richardson-2018-sentencepiece}. The maximum sentencepiece length is set to 16.
In particular, the final tokenizer is constructed by merging three small sub-tokenizers.
The first and largest sub-tokenizer is trained in Chinese, English, and 6 Southeast Asian languages (i.e., Indonesian, Malay, Portuguese, Tagalog, Thai, and Vietnamese), with a vocabulary size of 150k.
The second sub-tokenizer has a vocabulary size of 20k and covers nine languages, including Dutch, French, German, Greek, Italian, Polish, Russian, Spanish, and Ukrainian, to provide comprehension capabilities for these languages.
The third sub-tokenizer with a vocabulary size of 50k is trained exclusively on Chinese, which is used to further improve the performance of the model on Chinese.
After deduplication, these three sub-tokenizers form the final tokenizer with a vocabulary size of 180k.
For training, we sampled 2M pieces of data per language, resulting in a training set of 31.4M in size.
The vocabulary sizes of the three sub-tokenizers are determined based on our data distribution and data quality.

\begin{figure}[htbp]
    \centering
    \includegraphics[width=0.85\textwidth]{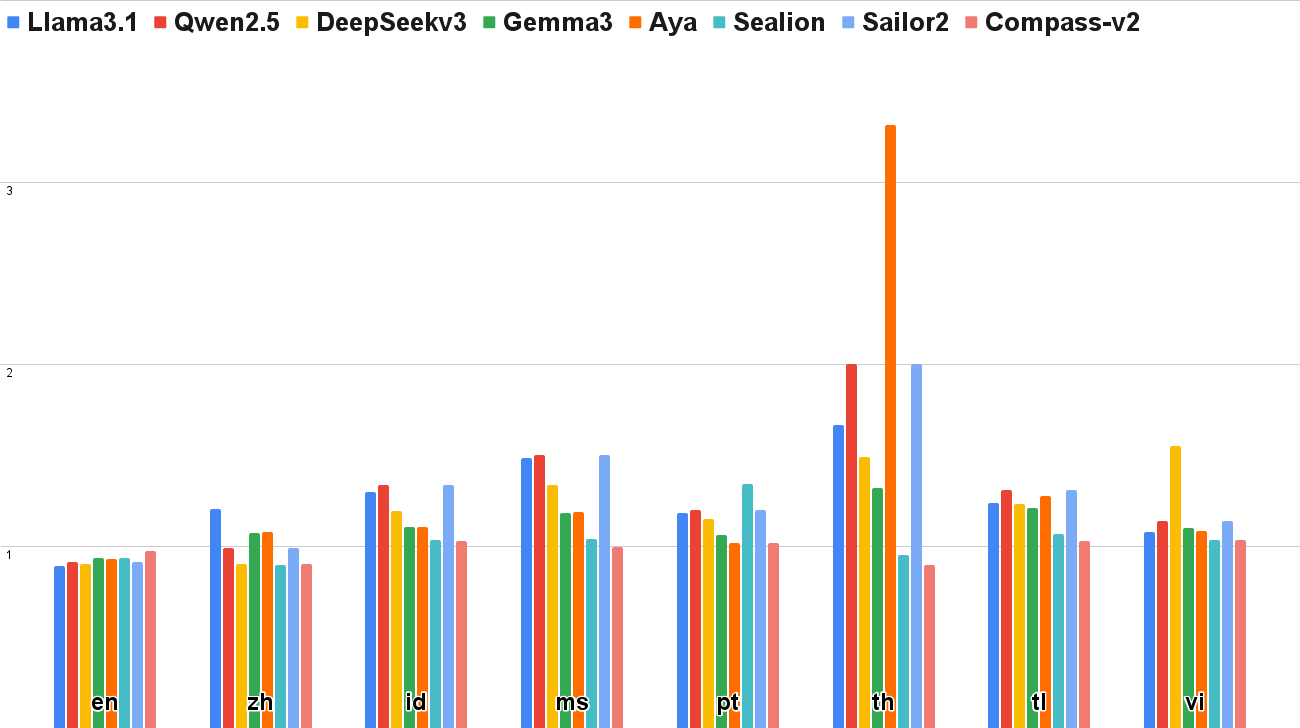}
    \caption{Compression Ratio on different languages of Compass-v2. Compass-v2 tokenizer achieves optimal compression rate for SEA languages.}
    \label{fig:compression rate}
\end{figure}

\textbf{Compass-v2 tokenizer achieves optimal compression rate for SEA languages.} To further evaluate the quality of the tokenizer, we measured its compression ratio (lower is better) across multiple languages and compared it with tokenizers from mainstream large models \citep{Dubey2024TheL3, Yang2024Qwen25TR, DeepSeekAI2024DeepSeekV3TR, team2025gemma} as well as those specialized for multi languages \citep{aryabumi2024aya, sea_lion_2024, sailor2report}.
The evaluation results are illustrated in Figure \ref{fig:compression rate}.
Compass-v2 tokenizer achieved the lowest values across all six Southeast Asian languages, particularly in Malay and Thai.
It effectively compresses multiple Southeast Asian languages and delivers balanced compression performance.
This proves our tokenizer to be the best-in-class solution for Southeast Asian language segmentation.

\subsubsection{Large industry-scale dataset size}
\begin{figure}[htbp]
    \centering
    \includegraphics[width=0.6\textwidth]{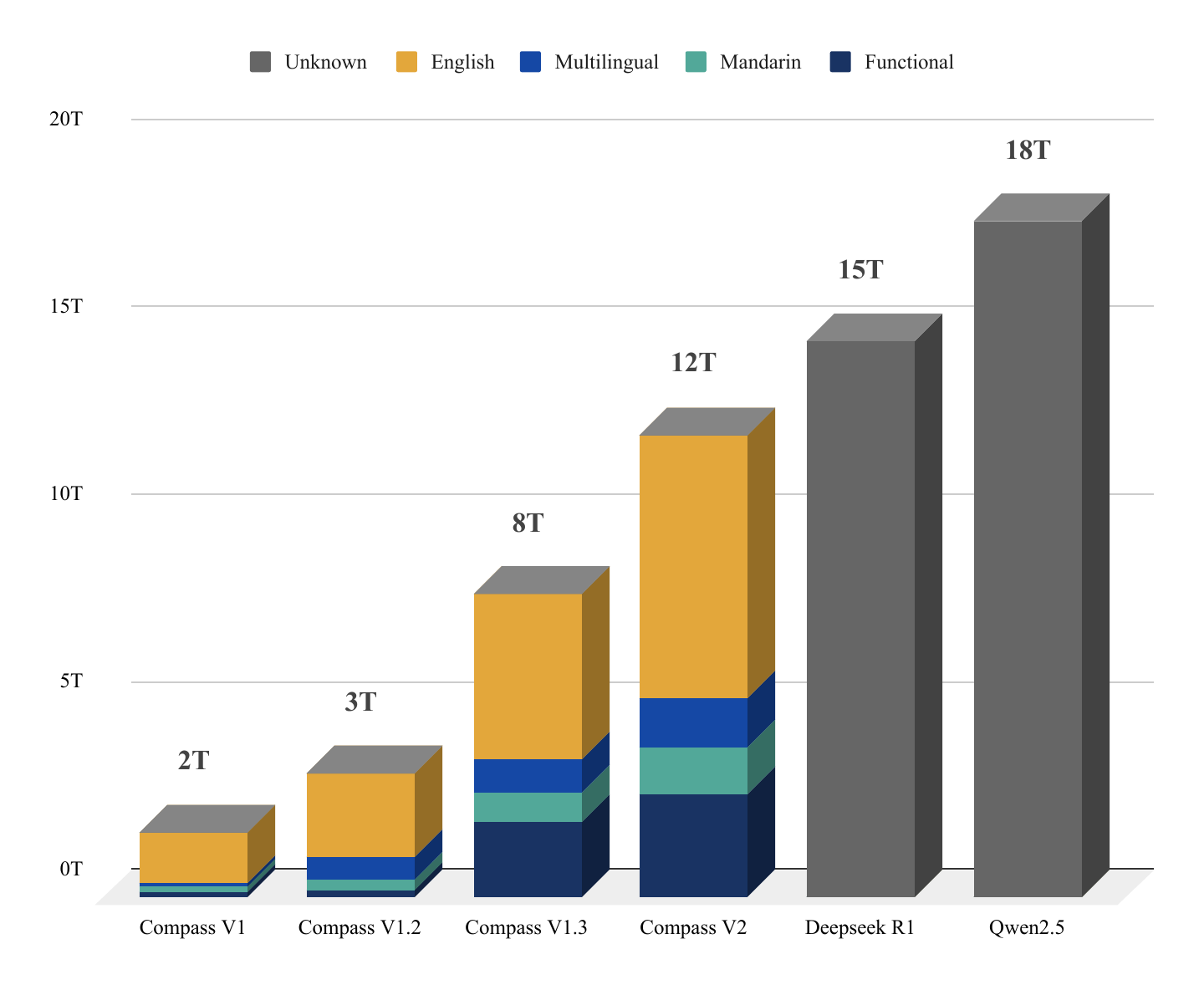}
    \caption{Data volume of Compass-v2 dataset, versus previous generations of Compass training data and leading models}
    \label{fig:data_volumn_compass}
\end{figure}

\textbf{Compass-v2 pretraining dataset has now reached a scale comparable to leading open-source and industry models} As illustrated in Figure~\ref{fig:data_volumn_compass}, the size of our pretraining dataset has been steadily increasing through iterative data collection and refinement. We increased our pretraining data corpus into 12 trillion tokens, which is more than 7-fold increase from the original dataset used in our first-generation Compass model pretraining~\citep{Maria2024CompassLM}. This significant expansion enables Compass-v2 to capture a broader range of linguistic patterns and domain knowledge. As shown in Figure~\ref{fig:data_volumn_compass}, our total data volume is now comparable to that of leading open-source models in the industry (such as Qwen2.5 and Deepseek-v3), positioning Compass-v2 at the frontier of large-scale language modeling. This growth not only improves coverage across languages and domains, but also enhances the model's ability to generalize from diverse and high-quality data sources.

\subsection{Architecture}

\begin{figure}[htbp]
    \centering
    \includegraphics[width=1\textwidth]{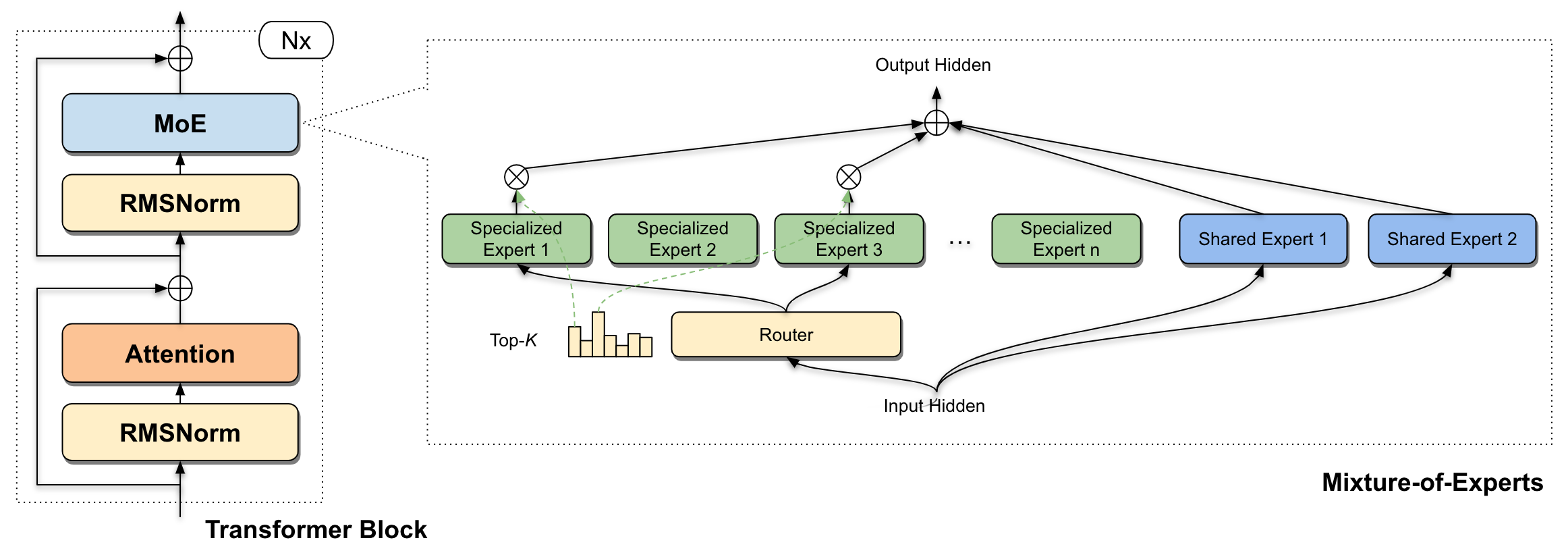}
    \caption{Model Architecture of Compass-v2. Compass-v2 applies the fine-grained expert design and it combines shared experts with specialized experts.}
    \label{fig:architecture}
\end{figure}

Compass-v2 comprises 5B activated parameters and 30B total parameters.
Starting from October 2024, we completed the preliminary model architecture design and refined each module through extensive small-scale experiments.
The model adopts a standard decoder-only MoE paradigm, as shown in Figure \ref{fig:architecture}, where traditional feed-forward network in transformer blocks are replaced by multiple experts and a routing layer.

Specifically, we incorporate the following key components:
\begin{itemize}[leftmargin=20pt, itemindent=-5pt]
  \item \textbf{Shared and Specialized Experts}: Compass-v2 combines shared experts with specialized experts to decouple knowledge at different granularities. Each token is assigned to 2 shared experts and 4 specialized experts, allowing shared experts to capture universal underlying knowledge, while specialized experts focus on task-specific complexities. \cite{sun2024hunyuan, liu2024deepseek, qwen2.5} have also validated the superiority of this strategy in model performance and downstream tasks. These updates of the model architecture also enhance the training stability and generalization capabilities of the model.
  \item \textbf{Fine-Grained Expert Design}: Since Compass-v2 is targeted at multiple different tasks in multiple languages, we use a smaller expert intermediate size, in other words, a fine-grained expert design. The inclusion of 2 shared experts boosts the model's nonlinear expressive power and general knowledge memory capability, while the fine-grained specialized experts enables richer expert combinations. This architecture significantly improves performance in multilingual, multi-domain and multi-task scenarios.
  \item \textbf{Dynamic Load Balancing}: A key challenge in training MoE models is how to ensure balanced token routing across experts, as skewed routing distributions can lead to underutilized experts or router collapse. To address this, we explicitly enforced load balancing through an auxiliary loss \citep{fedus2022switch} to penalize uneven expert usage, 
  $$\mathcal{L}_{aux} = N \cdot \sum\limits_{i=1}^{N} (\frac{p_i}{B} \cdot \frac{c_i}{B \cdot K})$$
  where $N$ is the number of experts, $p_i$ and $c_i$ are the aggregated routing probability and the activation count of expert $i$, respectively. $B$ is the number of tokens in one batch, and $K$ is the number of activated experts for each token.
  A Z-loss \citep{zoph2022st} is also used to stabilize router logits,
  $$\mathcal{L}_{Z} = \frac{1}{B} \sum\limits_{j=1}^{B}(\mathrm{log}\sum\limits_{i=1}^{N}\mathrm{exp}(z_i^j))^2$$
  where $z_i^j$ is the routing logits of expert $i$ on token $j$.
  To maintain optimal model performance, we disabled the droptoken mechanism during training. Additionally, we observed that the router's gating layer requires high numerical precision, so we switched its computation from bf16 to fp32 to prevent overflow issues.
\end{itemize}

The overall optimization objective for Compass-v2, incorporating auxiliary loss and Z-loss, is given by:
$$\mathcal{L} = \mathcal{L}_{LM} + \alpha \cdot \mathcal{L}_{aux} + \beta \cdot \mathcal{L}_{Z}$$
where $\alpha$ is the coefficient for auxiliary loss, $\beta$ is the coefficient for Z-loss. 
Furthermore, to achieve more stable model convergence, both coefficients $\alpha$ and $\beta$ are designed to progressively decay throughout the training process as the number of optimization steps increases.

\begin{table}[htbp]
\centering
\caption{Hyperparameters of Compass-v2}
\label{tab:hyperparameters}
\begin{tabular}{lc}
\toprule
\textbf{Configuration} & \textbf{Compass-v2} \\
\midrule
\# Layers & 30 \\
\# Attention Heads & 24 \\
\# Key/Value Heads & 6 \\
\# Shared Experts & 2 \\
\# Specialized Experts & 48 \\
\# Activated Specialized Experts & 4 \\
\# Attention Head Dimension & 128 \\
\# Expert Intermediate Size & 2048 \\
Position Embedding & RoPE \\
Activation Function & SwiGLU \\
\bottomrule
\end{tabular}
\end{table}

Table \ref{tab:hyperparameters} summarizes the key configurations of Compass-v2. 
We use Grouped Query Attention (GQA) \citep{ainslie-etal-2023-gqa} to maintain high performance and efficient memory utilization, along with Rotary Position Embedding (RoPE) \citep{10.1016/j.neucom.2023.127063} for position information modeling.
SwiGLU \citep{10.5555/3305381.3305478} is used as the activation function and RMSNorm \citep{zhang2019root} is used for hidden pre-normalization.
The initial context length of the model is set to 4096, which is extended to 32k in later training phase.

\subsection{Training}

We divide the pre-training process into three distinct stages, each serving a specific purpose in model development. These include: (1) an initial pre-training stage for broad knowledge acquisition, (2) an annealing stage to refine performance using high-quality data, and (3) a long-context expansion stage to extend the model's context window and enhance its long-sequence reasoning capabilities.


\subsubsection{First Stage: Large-Scale Pre-training}
In the initial pre-training phase, we used \textbf{8T of medium-quality multilingual data} to train the model, enabling the model to acquire multilingual and multi-domain knowledge from a variety of different data sources.

We adopted the AdamW \citep{loshchilov2017decoupled} optimizer with $\beta_1=0.9$, $\beta_2=0.95$ and a weight\_decay of 0.1. 
We follows a cosine learning rate decay schedule, with 3000 warmup steps, a maximum learning rate of $3.0 \times 10^{-4}$, and a minimum learning rate of $3.0 \times 10^{-5}$.
Gradient clipping was applied with a norm threshold of 1.0, and the global batch size was set to 5760.
The maximum coefficient of the auxiliary loss is set to 0.01 and the maximum coefficient of the Z-loss is set to 0.001.
For distributed training, we employed pipeline parallelism with a pipeline model parallel size of 4.

\subsubsection{Second Stage: High-Quality Annealing}
Following initial pre-training, we performed annealing using \textbf{4T tokens of high-quality data} to ensure stable training and sustained performance improvement during the final convergence stage.

During the annealing phase, we re-warmup the model for 2000 steps with learning rates ranging from $1.5 \times 10^{-4}$ to $3.0 \times 10^{-5}$.
This strategic re-warmup serves two key purposes: 1) it helps the model escape potential local optima , 2) it accelerates convergence on the high-quality data.
To further enhance the training stability, we increased the global batch size to $8640$ during this phase.

\subsubsection{Third Stage: Long-Context Extension}

\begin{figure}[htbp]
    \centering
    \includegraphics[width=1\textwidth]{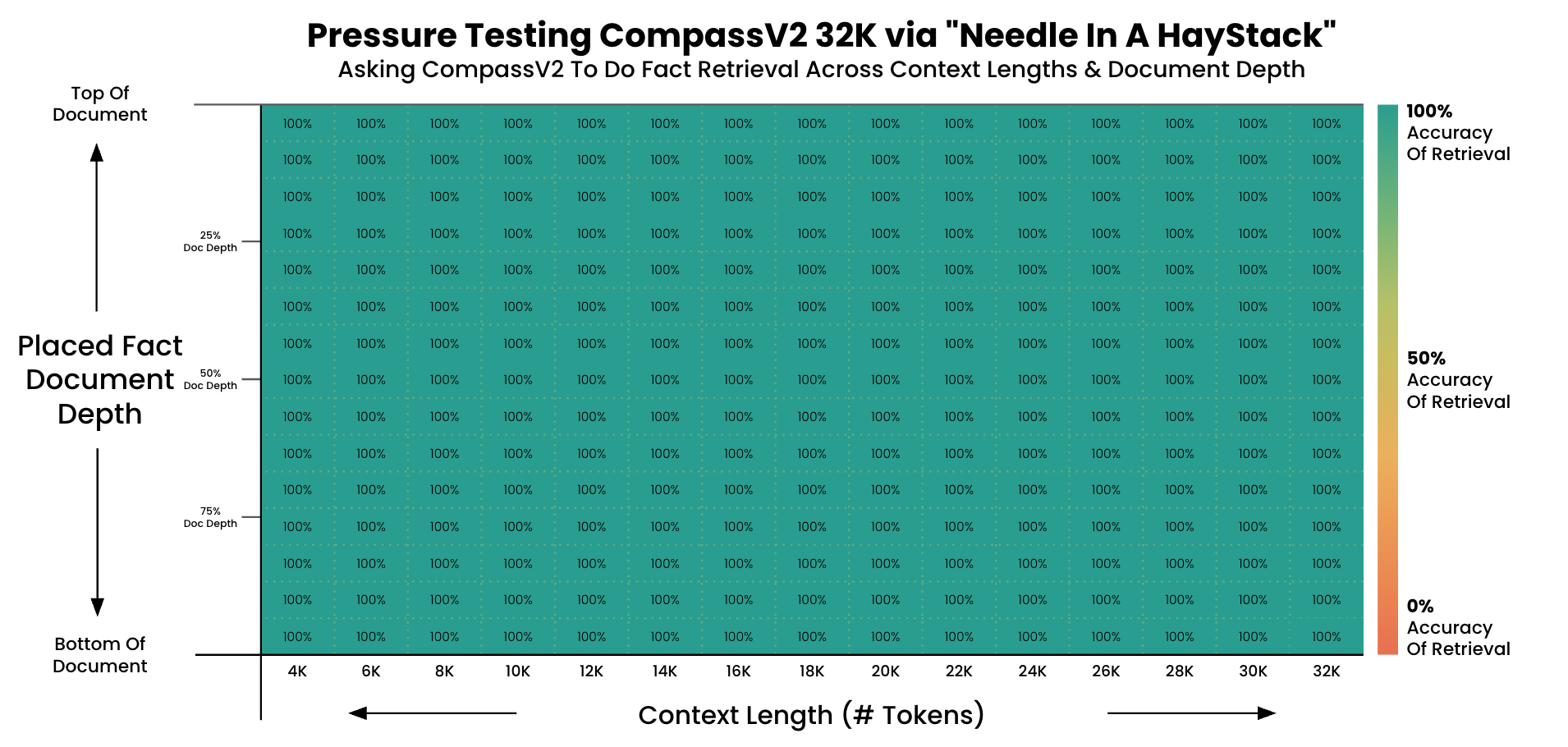}
    \caption{``Needle In A HayStack'' Performance of Compass-v2.}
    \label{fig:niah}
\end{figure}

In the third stage, we extend the context length of the model from $4096$ to $32768$, enabling the model to handle long texts. After pretraining the model on a 4k context window with a base frequency of RoPE of $10,000$, we increase the base frequency of RoPE to $1,000,000$ using the ABF techniques \citep{xiong2023effective} when we extend the context window to $32768$. The training dataset has the same distribution as the annealing dataset but is smaller; roughly $40\%$ of the data is under $4k$, and $60\%$ of it is between $4k$ to $32k$. 
We set the learning rate to $3.0 \times 10^{-5}$ and use a batch size of $2160$. 
Unlike the distributed strategy before training, we use \textit{context parallel superposition pipeline parallelism} to complete the long-text expansion training.
During continue pre-training, to prevent the degradation of general abilities, we keep monitoring the performance of the base model. We train until the model has satisfactory accuracy under various long-context metrics, such as the ``Needle In A Haystack'' and the line retrieval test. 
In the end, we continued pre-training the model for $2200$ steps. 
The trained model has \textit{strong long-context understanding ability while maintaining general ability}. Figure~\ref{fig:niah} shows it received perfect accuracy in the ``Needle In A HayStack'' evaluation.

Training Compass for long-text expansion requires a distributed parallel approach that differs significantly from traditional methods. To efficiently handle the computational challenges, we adopt a \textbf{hybrid parallelism strategy}, combining \textbf{context parallelism} and \textbf{pipeline parallelism}, which allows for more effective memory management and workload distribution.

A major challenge in long-text expansion is its potential negative impact on short-text quality. Conventional approaches often struggle to balance the two, but our method effectively preserves the model’s short-text generation capabilities while enhancing its ability to process longer contexts. 
We implement an \textbf{iterative training and evaluation strategy}, continuously assessing the model during training. The process continues until the model achieves \textbf{near-optimal performance} in generating \textbf{32K-token sequences}, while maintaining high accuracy in the target domain. This ensures that the model can handle both long and short text generation efficiently, making it well-suited for practical applications.

\section{Supervised Fine-tuning}\label{chapter:posttraining}
During the supervised fine-tuning (SFT) phase, we introduced a series of enhancements across both instruction design and training methodologies. On the instruction side, we focused on constructing high-quality multilingual and e-commerce datasets, and proposed a novel MOA method to enhance instruction quality. On the training side, we optimized the Megatron-LM framework to better support SFT process, and conducted a detailed analysis of various training configurations and their impact on model performance. 
\subsection{Supervised Fine-tuning Instruction}

\begin{figure}[h]
    \centering
    \begin{minipage}{0.98\textwidth}
        \centering
        \includegraphics[width=\textwidth]{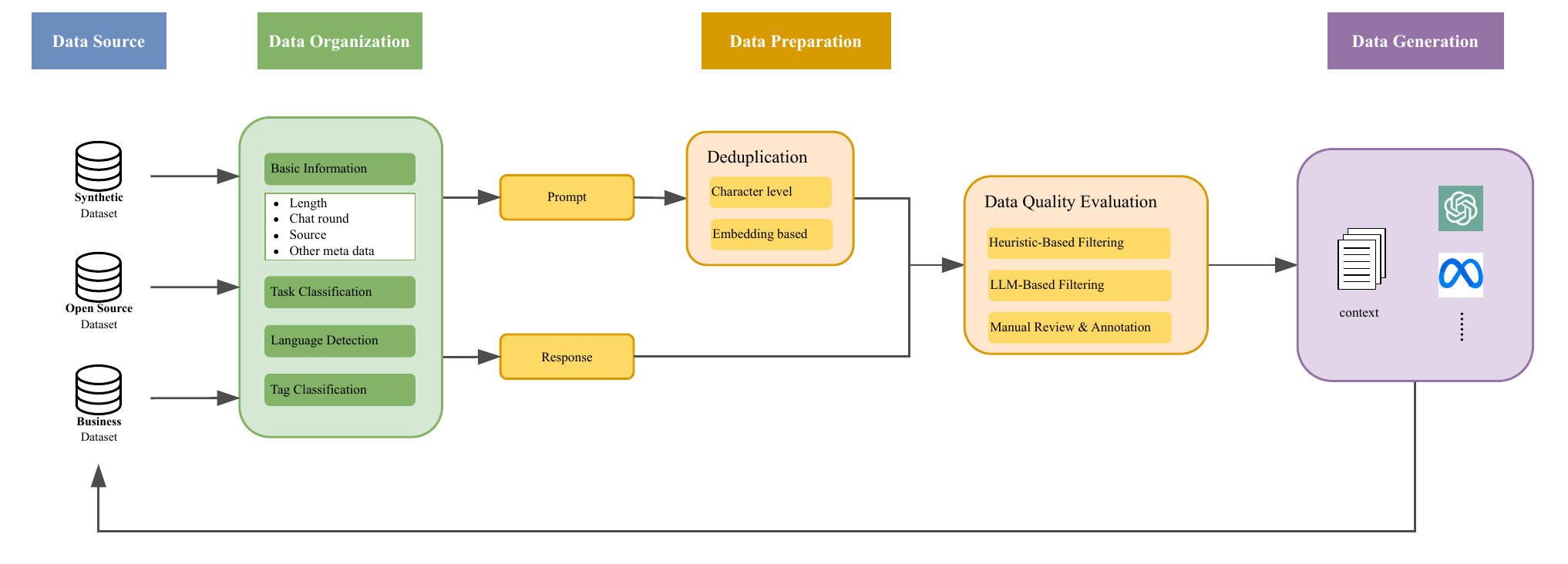}
        \caption{Overview of SFT data process pipeline}
        \label{figure: sft data processing}
    \end{minipage}
\end{figure}

\textbf{The current SFT instructions encompass approximately 6 million high-quality samples, encompassing the necessary SEA languages and a variety of e-commerce tasks within the industry.} Compass-v2 SFT dataset is sourced from three channels: open source data, synthetic data, and real business data.
To support robust multi-task capabilities, the instructions are organized at the task level into ten primary categories: safety, knowledge, understanding, mathematics, reasoning, creative generation, coding, e-commerce, agent, and platform-related information. These instructions are further subdivided into 70 subcategories, facilitating enhanced coverage across diverse use cases.
Given the model's deployment in the multilingual Southeast Asian market, the SFT dataset encompasses a diverse array of languages, including English (en), Chinese (zh), Tagalog (tl), Malay (my), Indonesian (id), Portuguese (pt), Vietnamese (vi), and Thai (th). 
To ensure precise language labeling, sentence-level language identification is applied instead of response-level identification. This approach allows for the more effective filtering of noisy translations and synthetic responses, thereby enhancing the overall quality of the SFT dataset. An overview of the SFT data processing pipeline is depicted in Figure \ref{figure: sft data processing}.

In the following sections, we provide detailed explanations and examples for each category of our SFT data.

\subsubsection{Multilingual Instructions} 
\textbf{Significant breakthroughs have been achieved in the development of multilingual capabilities, systematically enhancing the model's comprehension and generation abilities in complex multilingual scenarios.}
Building high-quality multilingual data pairs is crucial for global e-commerce platforms like Shopee. The core challenges lie in addressing 1) the scarcity of low-resource language data (e.g., Thai, Vietnamese), 2) the prevalent code-mixing phenomenon in user-generated content, and 3) the noise and heterogeneity in cross-lingual product information. These difficulties lead to significantly inferior performance of existing models on low-resource languages compared to dominant languages (e.g., English), directly impacting the accuracy of core functionalities like search recommendation and product management, as well as overall user experience. Through systematic multilingual data construction, we not only enhance model performance for low-resource languages and achieve cross-lingual knowledge transfer to reduce localization costs, but also support the platform's sustainable development in multicultural markets, balancing technical performance with linguistic inclusivity.

Facing Southeast Asia's unique linguistic diversity challenges, our solution demonstrates qualitative improvements across key dimensions including low-resource language understanding, code-mixing processing, and cross-lingual transfer. Particularly, while maintaining performance in dominant languages, we have achieved remarkable progress in low-resource languages and mixed-language scenarios, leading to more balanced service quality across language versions. These capability enhancements provide more accurate multilingual support for Shopee's core business scenarios including search recommendation, product management, and user interaction, effectively facilitating cross-border operations and optimizing localization experiences. This advancement not only represents a technological breakthrough but also establishes new benchmarks for multilingual services in global e-commerce platforms.

A multilingual dataset has been constructed, comprising three primary categories: 1) open-source instruction datasets (e.g., xP3~\citep{muennighoff2023crosslingual}, xP3mt~\citep{muennighoff2023crosslingual}, Alpaca-GPT4-Portuguese~\citep{Chen_MultilingualSIFT_Multilingual_Supervised_2023}, COIG~\citep{zhang2023chinese}, Vietnamese-395k-meta-math-MetaMathQA-gg-translated), 2) traditional NLP benchmarks (e.g., SQuAD~\citep{rajpurkar2016squad}, MS MARCO~\citep{DBLP:journals/corr/NguyenRSGTMD16}, TriviaQA~\citep{joshi-etal-2017-triviaqa}, HotpotQA~\citep{yang2018hotpotqa}, RACE~\citep{lai-etal-2017-race}, NewsQA~\citep{trischler-etal-2017-newsqa}, CoQA~\citep{reddy-etal-2019-coqa}), and 3) multilingual data obtained by translating selected subsets of these NLP benchmarks into multiple target languages. Open-source instruction data enhances the model’s ability to generalize instruction-following tasks across languages, while traditional NLP benchmarks strengthen fundamental language comprehension and reasoning skills. The translated data further promotes cross-lingual transfer and linguistic diversity, supporting robust multilingual instruction-following performance.

To align with Shopee’s application scenarios in Southeast Asia (SEA), the dataset emphasizes SEA languages and maintains a balanced linguistic distribution across Thai, Portuguese, English, Indonesian, Malay, Vietnamese, and Chinese. Additionally, we standardize heterogeneous raw data into uniform generative and multiple-choice templates, facilitating efficient training and ensuring compatibility across diverse data sources. To further refine the dataset, we apply heuristic filtering to remove irrelevant tasks (e.g., event linking) and reduce the proportion of overly simplistic tasks (e.g., text simplification), improving data relevance.

Since recent studies have shown that high-resource data, such as English, plays a crucial role in enhancing multilingual language models by improving their reasoning, comprehension, and generalization across low-resource languages~\citep{wang2025multilingual}, we incorporate representative subsets from traditional NLP datasets explicitly tailored for generative tasks.

Next, Our multilingual data selection pipeline begins with embedding generation using BGE-M3~\citep{chen-etal-2024-m3} for training samples across various languages. These embeddings are processed through PCA for dimensionality reduction followed by DBSCAN clustering. To optimize selection, we account for instruction diversity and question-answer alignment through two key operations: question clustering and Q/A similarity computation. The final sampling strategy incorporates both 'fuzzy' category labels and inversed Q/A scores, maintaining sample diversity while ensuring a balanced difficulty distribution in the selected training set. As a reuslt, we improve task diversity while achieving approximately an 83\% reduction in data volume without compromising key multilingual performance indicators, such as cross-lingual transferability and instruction-following accuracy.

\begin{figure}[htbp]

    \centering

    \includegraphics[width=\linewidth]{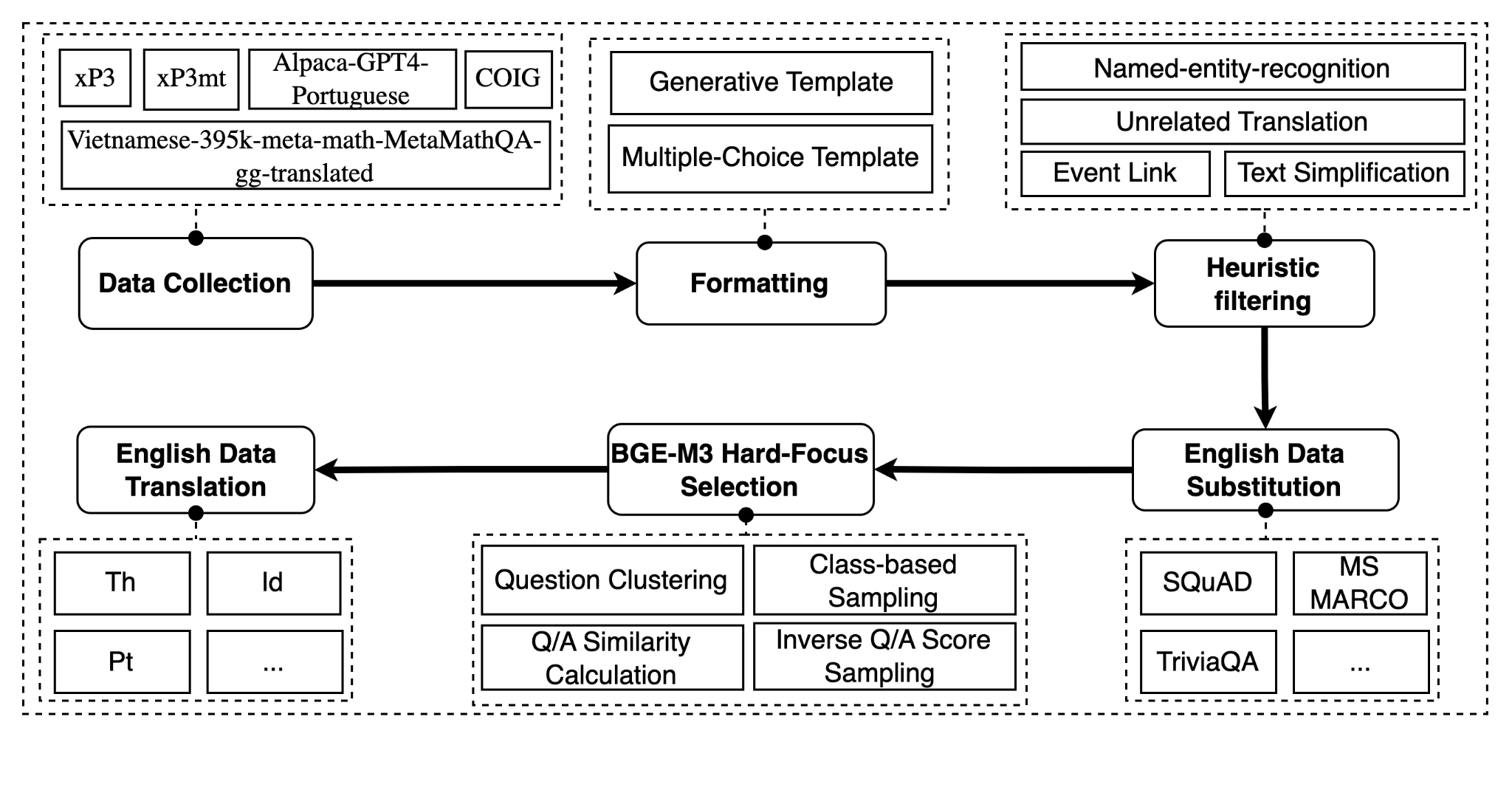}

    \caption{Overview of the multilingual dataset construction pipeline.}

    \label{fig:multilingual_pipeline}

\end{figure}

Finally, the curated high-resource English datasets are augmented into multiple languages using embedding-based techniques, effectively enhancing cross-lingual generalization and improving the model’s capability to understand and generate responses in diverse linguistic contexts. Modest but consistent improvements have been achieved across multiple standardized benchmarks, including Belebele for multilingual understanding, MMLUSea for Southeast Asian language evaluation, and MGSM for mathematical reasoning. Specifically, our model demonstrates performance gains of 10\%, 2.3\%, and 8\% on these benchmarks respectively. While these improvements appear numerically incremental, they represent meaningful progress in the model's capability to handle linguistically diverse and regionally specific tasks. An overview of the entire dataset construction pipeline is illustrated in Figure~\ref{fig:multilingual_pipeline}.

\subsubsection{E-Commerce Instruction Data}
\textbf{A comprehensive e-commerce instruction dataset has been developed, encompassing the widest range of e-commerce tasks. In our in-house evaluations, this dataset has demonstrated superior performance compared to Qwen2.5-14B and LLaMA3-70B, and it approaches the capabilities of GPT-4o.}

LLMs in the e-commerce domain have attracted widespread attention; however, most current LLMs lack specialized e-commerce instruction data for training. This deficiency often leads to generated responses that exhibit hallucinations and fail to properly understand product attributes, limiting their effectiveness in real-world e-commerce applications. To address these limitations and enhance LLM application effectiveness in the e-commerce domain, we have developed a comprehensive e-commerce instruction dataset. This dataset has been systematically designed to address specific e-commerce needs and improve model performance across various e-commerce tasks.

We have abstracted e-commerce domain capabilities into 12 major categories:
\begin{enumerate}
    \item \textbf{Product \& Brand Information}: Knowledge about products, brands, and their relationships
    \item \textbf{Attribute Extraction}: Identifying and extracting key product attributes from descriptions
    \item \textbf{Product Classification}: Categorizing products into appropriate taxonomies
    \item \textbf{Product Recommendation}: Suggesting relevant products based on user preferences
    \item \textbf{Product Similarity Judgment}: Assessing similarity between different products
    \item \textbf{Numerical Reasoning}: Performing calculations related to pricing, discounts, item quantities and inventory
    \item \textbf{Product Information Generation}: Generating product titles, descriptions, summaries, and other product-related content
    \item \textbf{Product Questioning}: Answering specific queries about product features and usage
    \item \textbf{Query-Product Matching}: Aligning user search queries with relevant products
    \item \textbf{Review Understanding}: Performing sentiment analysis on product reviews and customer feedback
    \item \textbf{Query Rewriting}: Reformulating user queries to improve search results
    \item \textbf{Dialogue Understanding}: Multi-round shopping conversation intent analysis, intelligent customer service response.
\end{enumerate}

\paragraph{Data Sources and Methodology}
The e-commerce data is sourced from diverse real-world datasets including search engines, AmazonM2 \citep{jin2023amazonm2multilingualmultilocaleshopping}, Shopee product information, and other open and closed-source datasets. This diversity ensures broad coverage across different product categories and market segments.

\begin{figure}[t]
    \centering
    \includegraphics[width=\linewidth]{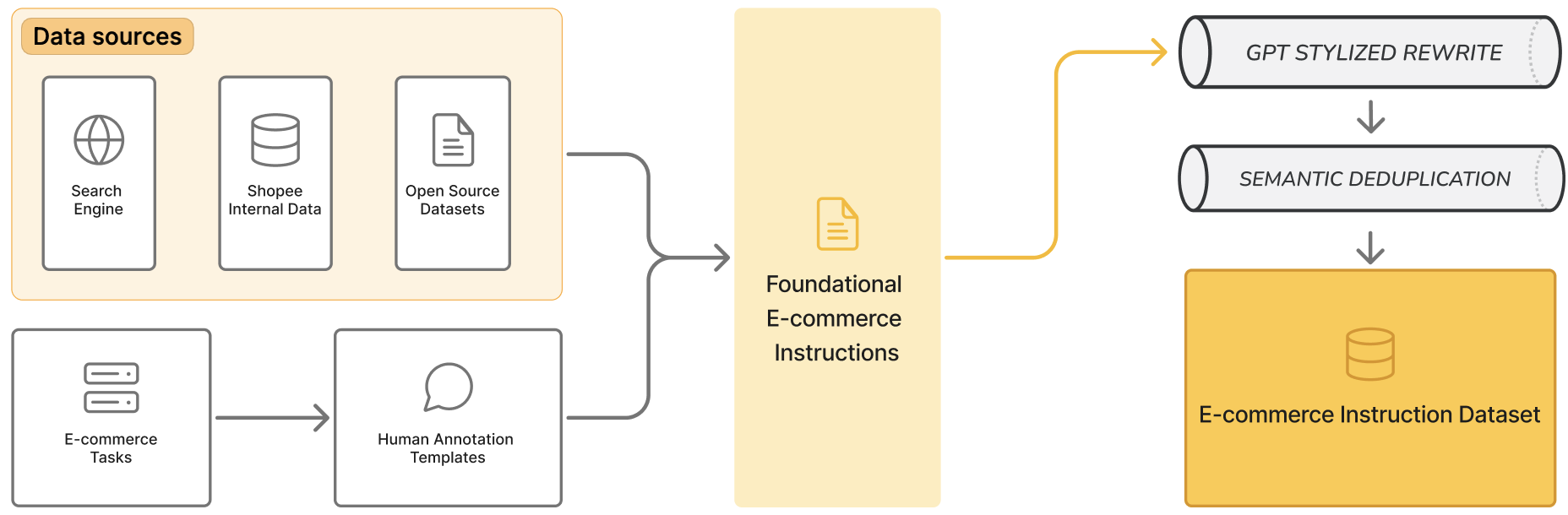}
    \caption{Overview of the e-commerce dataset construction pipeline.}
    \label{fig:e-commerce_pipeline}
\end{figure}

To construct high-quality instruction data, we employed a hybrid methodology combining manual annotation and GPT-4o \citep{openai2024gpt4technicalreport} reformulation, as illustrated in Figure~\ref{fig:e-commerce_pipeline}.
\begin{itemize}
    \item For each e-commerce task, we developed various instruction templates through human authorship.
    \item Integrate the templates with real e-commerce data to ensure the incorporation of e-commerce domain knowledge and generate foundational instructions.
    \item Multiple style-specific templates were established to guide GPT-4o in reformulating these instructions with diverse tones, perspectives, and complexity levels.
    \item To enhance reasoning capabilities, we incorporated Chain-of-Thought methodologies in responses for select tasks, particularly those requiring multi-step reasoning or complex product comparisons.
\end{itemize}

\paragraph{Quality Assurance}
Recognizing that homogeneous data may compromise a model's generalization capabilities, we implemented a rigorous similarity-based deduplication process. This process employed semantic embeddings to identify and remove excessively similar instructions, ensuring that the final dataset represents a diverse array of e-commerce scenarios and query types.

\begin{figure}[t]
    \centering
    \includegraphics[width=\linewidth]{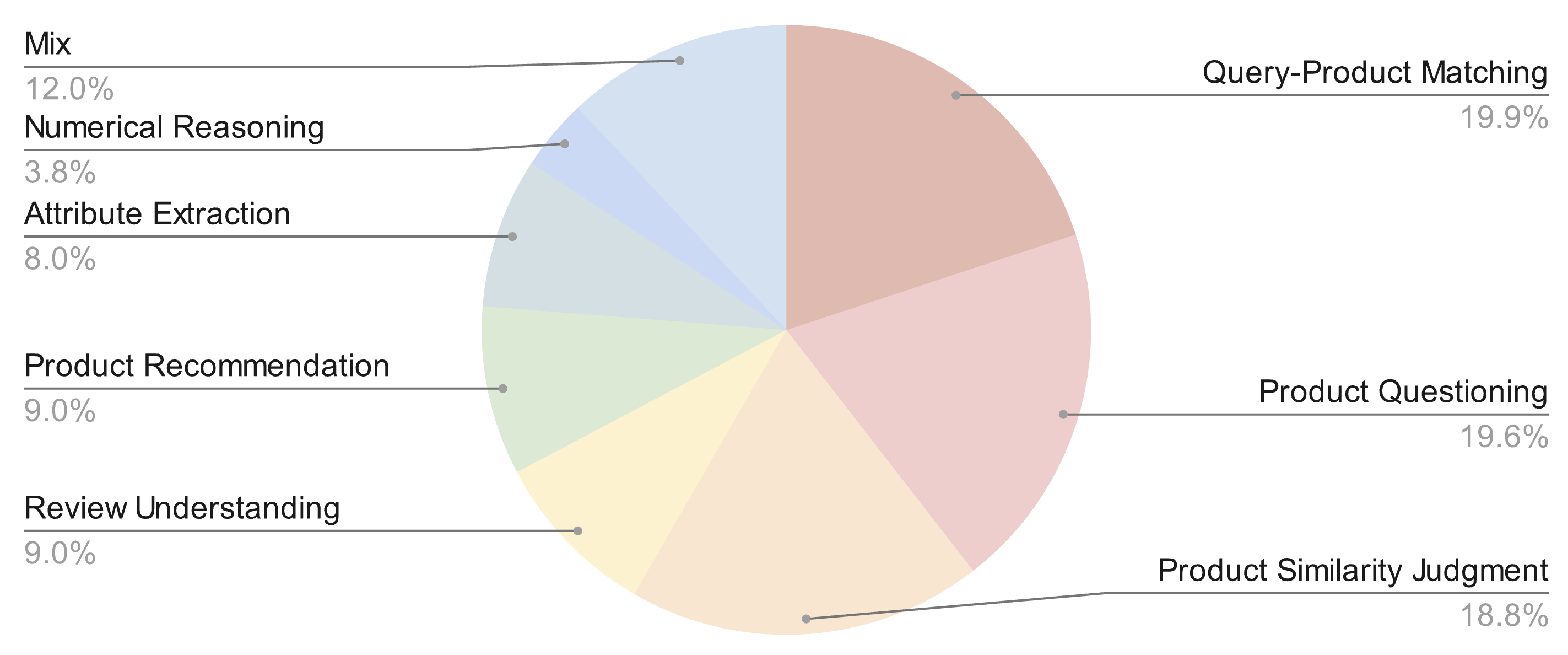}
    \caption{E-commerce Tasks Distribution.}
    \label{fig:e-commerce-task-distribution}
\end{figure}

\paragraph{E-commerce Tasks Distribution}
The Figure\ref{fig:e-commerce-task-distribution} illustrates the E-commerce data tasks distribution. The data is predominantly composed of three primary categories: Query-Product Matching, Product Questioning and Product Similarity Judgment, collectively representing over half of the total data. These data provide a substantial amount of query and product information, which is crucial for enhancing the model's capabilities in e-commerce applications. The data in the Review Understanding category provide information on user reviews from e-commerce platforms, which are an essential component of the e-commerce domain. Additionally, the data from Product Recommendation and Attribute Extraction enable the model to learn the requirements and paradigms of specific e-commerce tasks. The Numerical Reasoning category facilitates the transfer of the model’s reasoning capabilities to the e-commerce domain. Given the model’s inherently strong reasoning ability, only a limited amount of additional data is required. The Mixed category consists of a combination of data from other multiple tasks. Since the data corresponding to each individual task are relatively sparse, they are aggregated and presented as "Mix" in the figure. This category primarily addresses the variations in format and content requirements across different tasks.

\subsubsection{Other Data} 

\textbf{Function Calling Data:} We curate a function calling data of 154{,}563 raw prompt-response pairs, each optionally annotated with a \texttt{tools} field containing structured metadata on tool usage. To support supervised training and evaluation, we prioritize entries with non-empty \texttt{tools} fields, selecting for diversity in tool schemas and argument complexity to ensure broad coverage and improve generalization. 

\textbf{Question Answering Data:} We curate a high-quality question answering dataset derived from HotpotQA~\citep{yang2018hotpotqa}, systematically reformulating instances along four key dimensions inspired by~\citep{chen2023benchmarking}: contextual noise resilience, negative rejection, information integration, and counterfactual stability.

\textbf{Coding Data:} To enhance the model's coding capabilities, we construct an instruction-tuning coding corpus encompassing nearly 40 programming languages by integrating open-source data with domain-specific synthetic data. Our dataset covers a broad spectrum of topics and instruction types, ranging from algorithmic problem-solving to package-specific knowledge, and includes tasks such as conventional Q\&A, code completion, test case generation, and code refinement. 

\textbf{Role-playing Data:} We construct a role-playing dataset by combining open-domain corpora~\citep{li2023chatharuhirevivinganimecharacter, wang2025coser} with synthetic dialogues generated via instruction-tuned LLMs. To broaden representational diversity, the dataset integrates two complementary dialogue settings: authentic conversations between real-world characters extracted from books and fictional interactions involving synthetic personas in imagined scenarios. 


\subsection{Training} \label{chapter:sft}
\subsubsection{Two-Stage Methodology}
In our large-scale Supervised Fine-Tuning (SFT), we designed a two-stage training strategy to develop a model that not only follows general instructions but also demonstrates deep reasoning abilities and strong e-commerce and multilingual capabilities. Balancing these aspects posed significant challenges, requiring careful data selection and curriculum design.

In the \textbf{first stage}, we focused on establishing strong \emph{generalization and instruction-following capabilities}. The model was fine-tuned on a diverse dataset covering various domains, including information processing, commonsense QA, text generation, coding, logical reasoning, STEM, and safety-related tasks. This foundational phase enables the model to understand and execute a broad range of instructions effectively.

In the \textbf{second stage}, we introduced additional targeted datasets to refine \emph{reasoning abilities and domain expertise}:

\begin{itemize}
    \item \textbf{Mathematical reasoning data}: We incorporated problem-solving datasets with explicit thought processes wrapped in <think></think> tags, enhancing the model's ability to generate structured, multi-step solutions.

    \item \textbf{Selective resampling of high-quality data}: To maintain distributional alignment with the first stage, we sampled high-quality instruction data across multiple domains and integrated it into this phase.
    
    \item \textbf{Expanded reasoning steps}: A subset of the first-stage data, particularly multiple-choice questions (A/B/C/D), was enhanced with detailed step-by-step explanations to improve logical inference.
    
    \item \textbf{E-commerce data augmentation}: Specialized datasets focusing on product descriptions, customer inquiries, and transactional reasoning were added to strengthen the model’s ability in e-commerce-related tasks.
    
    \item \textbf{Role-playing and function-calling tasks}: To improve adaptability in interactive and API-driven applications, we incorporated data for role-based conversations and function execution.
\end{itemize}

Our results indicate that this mixed-training approach effectively combines general instruction-following with advanced reasoning capabilities. The model not only excels in complex mathematical reasoning tasks, but also demonstrates deeper logical thinking in general inference tasks. By structuring the training in this manner, we successfully developed a model capable of both general instruction execution and advanced multi-step reasoning, while also enhancing its performance in e-commerce and multilingual scenarios.

\subsubsection{Framework Optimization}
During pretraining, the primary focus is on learning from documents and their constituent tokens. However, in instruction fine-tuning, samples consist of instructions, questions, and corresponding answers. While Megatron-LM is inherently designed for pretraining scenarios, it cannot be directly applied to instruction fine-tuning. Additionally, in post-training, training samples need to maintain their integrity, but variable-length samples often lead to excessive padding tokens that negatively impact training efficiency. To address these challenges, we have enhanced Megatron-LM with the following key improvements for post-training support:

\begin{itemize}
    \item \textbf{Preprocessing Pipeline Optimization}: Redesigned the underlying preprocessing pipeline to ensure compatibility with post-training requirements
    
    \item \textbf{Sample Packing Strategy}: Introduced an efficient packing algorithm to concatenate multiple samples, improving both training speed and GPU utilization
    
    \item \textbf{Distributed Dynamic Padding (DDP)}: Developed a novel padding technique that maintains consistent input sizes across data parallel groups while maximizing valid token proportion
    
    \item \textbf{Communication Strategy Refinement}: Optimized framework communication patterns specifically for instruction fine-tuning scenarios
\end{itemize}

Specifically, we compared the performance of several frameworks capable of instruction fine-tuning, including Transformers, LLaMA-Factory, and OpenChat. While these frameworks meet basic post-training requirements, the substantial parameter size of the models and considerations for subsequent training efficiency reveal that using only the Data Parallelism (DP) mode in these frameworks introduces additional communication overhead and underutilizes GPU resources. Consequently, we developed a post-training framework based on Megatron-LM.

Our initial approach introduced a MASK tensor as additional input information to the framework, enabling implementation with minimal modifications. However, since post-training places greater emphasis on sample integrity and the variability of token counts per sample, this method results in excessive padding tokens that significantly reduced the proportion of effective computations.

Subsequently, we conducted the experiments with a dynamic padding strategy to ensure all input tokens were valid, but this approach caused a noticeable decrease in training speed (similar to the issues mentioned in the FSDP paper \cite{zhao2023pytorch}). We therefore designed a Distributed Dynamic Padding (DDP) strategy that maintains consistent input sizes across all DP groups (albeit with some efficiency trade-offs), achieving more satisfactory training speeds.

To further enhance training efficiency while maintaining sample integrity, training speed, stability, and the accuracy of our Megatron-LM-based post-training framework, we implemented sample packing. This technique enables processing more samples per forward/backward pass under the same batch size configuration. We also optimized the communication strategies of existing parallel approaches to accommodate this training paradigm. Comparative experiments demonstrate that our framework achieves several tens of percent faster performance than the aforementioned frameworks under similar configurations, while maintaining comparable accuracy with average step loss differences not exceeding three decimal places.

\section{Reinforcement Learning}\label{chapter:DPO}

\subsection{Reward modelling}
The Reward model (RM) was trained using an open-source foundation and open-source data. The foundation of the model is the pre-trained version of Qwen2.5-14B~\citep{Yang2024Qwen25TR,qwen2}. The data used is from the Preference-700k dataset~\citep{dong2024rlhf}. 
A total of four models were trained using different hyperparameter configurations, varying the number of epochs $[1, 2]$, learning rate $[0.000003, 0.000004]$ and learning rate schedules (linear or cosine).
Finally, the final RM model was obtained through weight averaging of these four models.

We conducted several comparison experiments and found that the publicly available leaderboard RewardBench suffers from limited data volume and inconsistent data quality, which makes it difficult to conduct reliable and robust evaluations.
For example, we found that although some RM models like ArmoRM perform well on Rewardbench, they showed poor results in our in-house testing. Hence, instend of using Rewardbench's pairwise accuracy, we chose the accuracy of Best-Of-N (BON) sampling as our in-house testing standard.
To objectively measure the accuracy of sample responses, the datasets used for BON evaluation primarily focus on mathematical abilities, including a verified GSM8K test set and a verified subset of the AQuA test set.
After considering both Rewardbench and in-house tests, we ultimately selected the RM model with the best performance.

\subsection{Online Reinforcement Learning}
Reinforcement Learning from Human Feedback (RLHF)~\citep{rlhf} has emerged as a powerful paradigm to align Large Language Models (LLMs) with human values and preferences. In the Post-Training phase of Compass-v2, we adopted the Online-Direct Preference Optimization (Online-DPO)~\citep{dpo,online_dpo,iterative_dpo} algorithm to align the large language model. Online-DPO, which has higher training efficiency and a more stable training process, have recently emerged as efficient alternatives to Reinforcement Learning from Human Feedback (RLHF). DPO~\citep{dpo} can directly optimizes the RLHF objective using the following equivalent formulation:
$$\mathcal{L}_{DPO}= -\mathbb{E}_{(x, y_{w}, y_{l})\sim D}
[log\sigma(\beta log\frac{\pi_\theta(y_{w}|x)}{\pi_{ref}(y_{w}|x)} - 
\beta log\frac{\pi_\theta(y_{l}|x)}{\pi_{ref}(y_{l}|x)})]$$

\begin{figure}[htbp]
  \centering
  \includegraphics[width=1.0\textwidth]{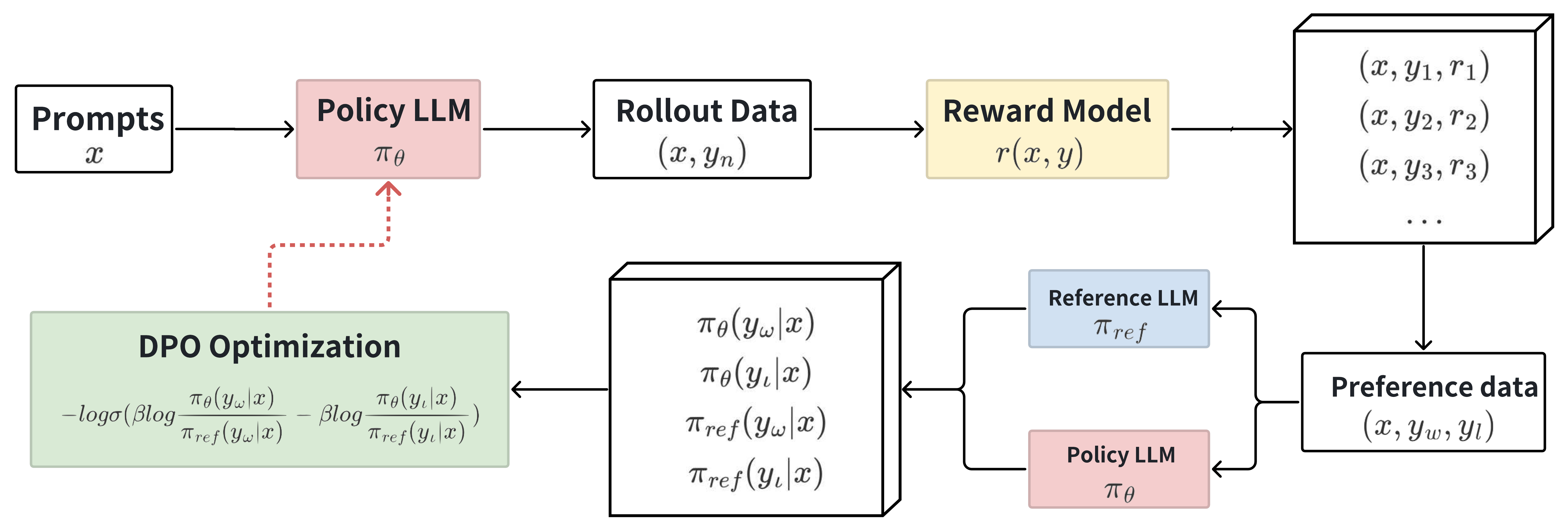}
  \caption{Overview of Online-DPO Pipeline in Compass-v2.}
  \label{fig:online-dpo}
\end{figure}

  \textbf{Prompts Data Construction}: We constructed a prompt-training dataset for the preference optimization phase from different sources, including general data, bad-case data, mathematical data, and e-commerce data:
\begin{enumerate}
    \item The general data comes mainly from the UltraFeedback \citep{cui2023ultrafeedback} dataset. In order to facilitate the adjustment of the data recipe, we carefully categorize the dataset, and form a dataset which mainly covers aspects such as creative generation, code, reasoning, mathematics, common sense, and security.

    \item 
    The bad-case data is sourced from user queries collected through our online services, primarily consisting of multilingual instruction-following tasks, identity recognition questions, and commonsense questions that are frequently misanswered by the model.

    \item The mathematical data comes from open source datasets, such as the gsm8k dataset \citep{cobbe2021gsm8k} in multiple languages, such as English, Chinese, Thai, etc. We use these data as a supplement to our general training corpus, aiming to enhance the model’s mathematical reasoning capabilities.

    \item The e-commerce data is a subset of our SFT e-commerce dataset. Based on the evaluation result of our SFT model, we specifically selected different types of e-commerce data as data supplements for the preference optimization stage.
\end{enumerate}

\textbf{Preference Data Construction}: The construction of preference data consists of two stages. The first stage is to generate answers using our Base Model. The second stage is to use the reward model to label the generated answers and filter the preference data based on the results and certain rules. The filtering rules are stated as follows:
\begin{enumerate}
    \item 
    During the answer generation phase, we leverage vLLM~\citep{vllm}, a fast and user-friendly library for large language model inference and serving, to accelerate response generation. For each prompt, we sample $n=32$ candidate answers. The setting of $n=32$ is determined empirically to balance generation efficiency and response diversity. To further enhance diversity, we use a relatively high temperature and apply different random seeds during generation, ensuring a wide range of outputs for downstream preference filtering.

    \item In the rewarding stage, we use the pre-trained reward model to annotate the generated answers, assigning a reward value to each response. 
    The strategy for selecting preference data varies based on the data type. 
    For general data, we rank the $n$ responses for each prompt according to their reward scores and select appropriate samples as the chosen and rejected answers by analyzing the reward distribution. For math data, we apply an additional Rule-Based Reward Function to annotate the answers. 
    Specifically, we parse the final answer from each response and compare it to the ground truth to determine whether the solution is correct. Based on this, we divide the $n$ answers to each math prompt into two groups: one correct group $G_T$ and one incorrect group $G_F$. If all answers to a question are incorrect, the question will be discarded. Finally, we select the answer with the highest reward in $G_T$ as the chosen answer and the answer with the highest reward in $G_F$ as the rejected answer to construct the math preference data.
\end{enumerate}

\section{Hybrid Reasoning Model}
\subsection{Motivation}
Instilling complex reasoning capability in LLMs has been recognized as a significant challenge. 
Recently, there has been an increasing emphasis on the long-chain reasoning capabilities of LLMs. Many models ~\citep{openaio1,guo2025deepseek,QwQ} have demonstrated significant potential in advancing tasks that require deep, multi-step reasoning. For example, OpenAI-o1 \citep{openaio1}, which pioneered the concept of inference-time scaling by extending the response length, has shown a remarkable ability to manage long chain-of-thought data, enabling more sophisticated problem-solving approaches. Similarly, DeepSeek-R1 \citep{guo2025deepseek} has also underscored the importance of such reasoning in various complex domains, offering promising improvements over previous models that struggled with maintaining coherence over extended chains of logic. 

However, while these o1-like reasoning-focused models excel at handling complex, multi-step reasoning tasks, they often struggle in the scenarios that requires flexible and general responses~\citep{zhao2025trade}. 
This limitation stems from their narrow optimization on reasoning abilities, which restricts their ability to handle unstructured outputs or broader language understanding. 
In some real-world scenarios where tasks dynamically vary between simple queries and complex reasoning, neither a purely reasoning-driven model nor a general-purpose model alone is sufficient to handle both tasks. This can lead to inconsistencies and performance bottlenecks, severely affecting the user experience.

\textbf{To address these issues, we propose Compass-v2 as a Hybrid-Reasoning-Model \citep{claude3.7}, which integrates both deep reasoning capabilities and broader task adaptability into one model.} By combining the strengths of strong reasoning with improved generalization, the model aims to strike a balance between complex inference and task versatility, thereby enhancing the overall performance across a wider range of applications. This hybrid approach promises to address the current gaps in reasoning systems, making them more robust for a variety of tasks.

\subsection{Data}

To enhance reasoning capabilities, we introduce a Long Chain-of-Thought data with thinking process and summarized answer. We began by collecting a diverse range of sources of math queries with groundtruth answers as seed set (e.g. NuminaMath \citep{numina_math_datasets}, NuminaMath-1.5 \citep{numina_math_datasets1.5}, LIMO \citep{ye2025limoreasoning}, S1K-1.1 \citep{muennighoff2025s1simpletesttimescaling}, etc.). We implement data deduplication with exact matching and thorough decontamination against MATH500, AIME24, AIME25 and GPQA. The data is then filtered for complexity, where we initially employ previous Compass-v2 model to roll out responses several times for collected queries. Math-Verify is applied to retrieve correct examples. Queries with a pass rate that falls within the interval [$\alpha$, $\beta$] are selected, where the process helped set an initial difficulty threshold to filter out exceptionally challenging or trivial problems. The retaining queries are used for obtaining DeepSeek-R1 \citep{guo2025deepseek} responses and then add system prompt for activation. 

\subsection{Training}

\begin{table}[h]
    \begin{tcolorbox}[title=LongCoT, colframe=red!50!black, colback=red!5!white]
    \textbf{<System>:} Let's think step by step. Please ensure that the reasoning process are enclosed within <think> </think> tags, i.e., <think> reasoning process here </think>. And put your final answer within \verb|\boxed{}|.
    
    \textbf{<Human>:} Sample Query
    \tcblower
    
    \textbf{<AI>:} <think> Sample Thinking Process </think> Sample Answer.
    \end{tcolorbox}
    \begin{tcolorbox}[title=General, colframe=blue!50!black, colback=blue!5!white]
    \textbf{<Human>:} Sample Query
    \tcblower
    
    \textbf{<AI>:} Sample Answer
    \end{tcolorbox}
    \caption{Responses difference between LongCoT and general chat templates. LongCoT chat template is applied to trigger the thinking process of Compass-v2. }
    \label{longcot chat template}
\end{table}

In our SFT training pipeline, we implemented a \emph{two-stage training paradigm} as mentioned in Chapter \ref{chapter:sft} to optimize model performance. 
Benefiting from incorporating mathematical training data in the first phase, our base model has established a strong foundation for reasoning capabilities, enabling more accurate and efficient problem-solving. Building on this foundation, we add Long-CoT data in the second stage of training for further improving the reasoning capability. 
However, we observe that excessive exposure to Long-CoT data can negatively impact the model’s general language capabilities, leading to an undesirable trade-off~\citep{zhao2025trade}.
To mitigate this seesaw effect, we adopt downsampling on Long-CoT data to carefully balance its integration to enhance reasoning capabilities while preserving the overall robustness of the model. As a result, the trained model fully possess the Long-CoT reasoning ability \emph{within a 1\% drop} in the general metrics. Table \ref{longcot chat template} shows the responses difference of reasoning model activation. 

\section{Inference}\label{chapter:inference}
For Compass-v2, we leverage AWQ and FP8 quantization techniques to enhance efficiency. We evaluate the inference performance of Compass-v2 against Compass-v1 and other competitors.

\begin{figure}[htbp]
    \centering
    \begin{minipage}{0.82\textwidth} 
        \centering
        \includegraphics[width=\textwidth]{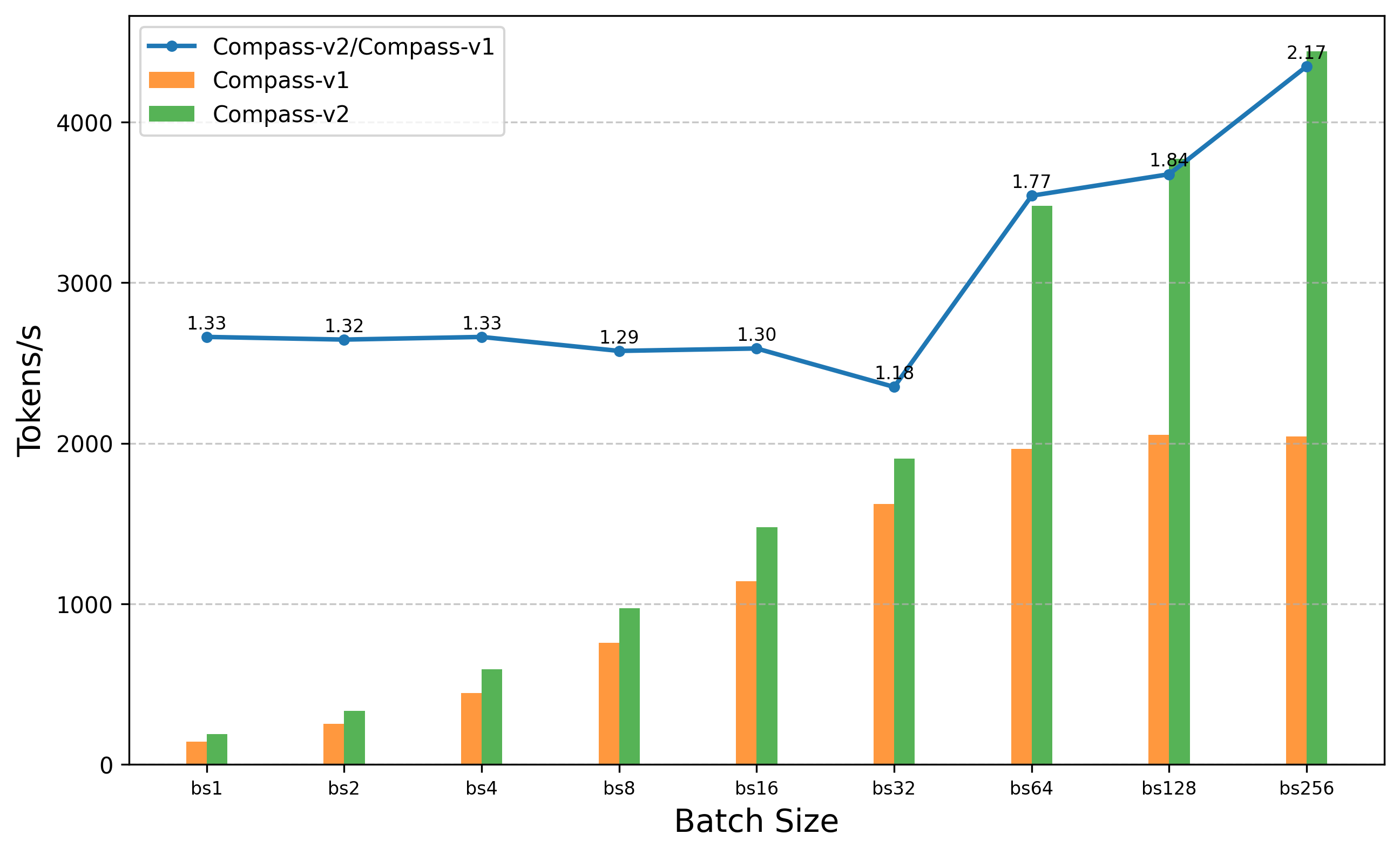}
        \caption{Comparison of inference speed between Compass-v2 and Compass-v1.}
        \label{fig:inference speed of Compass-v2 and Compass-v1}
    \end{minipage}    
\end{figure}

\subsection{Compass-v2 vs Compass-v1(13B)}

\textbf{Compass-v2 achieves an average speedup of 1.64× across various batch sizes compared to Compass-v1.} We compare the inference speed of Compass-v1 and Compass-v2 using in-house datasets derived from our chatbot service.  The average input length is approximately 1,200 tokens, with the model generating an average of 200 tokens per request. We send requests with concurrency levels from 1 to 256 and measure output throughput (tokens per second) as the total output tokens divided by processing time. As shown in Figure \ref{fig:inference speed of Compass-v2 and Compass-v1}, Compass-v2 exhibits superior performance at different concurrency levels compared to Compass-v1.

\subsection{Compass-v2 vs Competitors}

\begin{figure}[htbp]
    \centering
    \begin{minipage}{0.82\textwidth} 
        \centering
        \includegraphics[width=\textwidth]{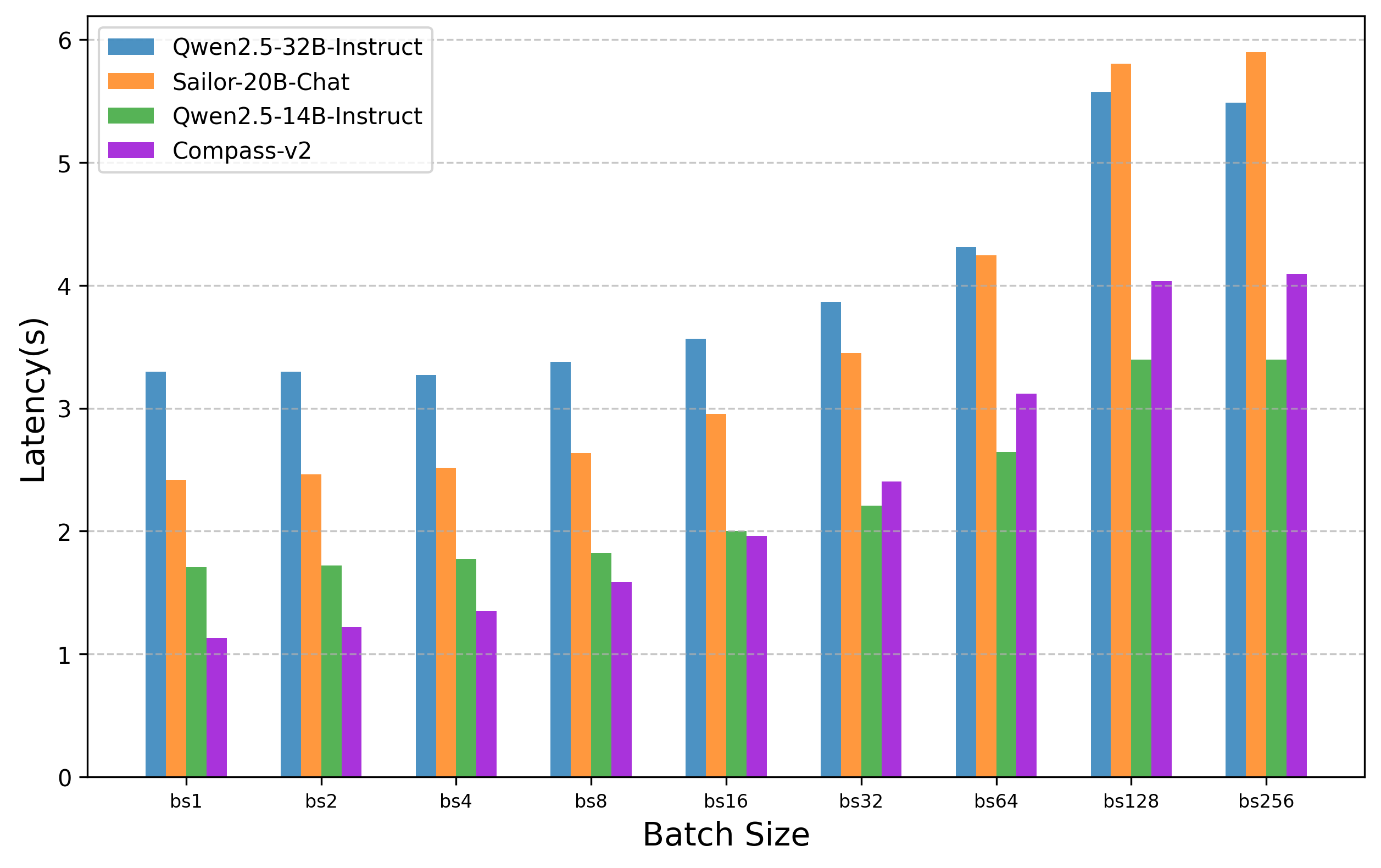}
        \caption{We compare the average inference latency across models of similar scale at various batch sizes. All competitors are deployed using state-of-the-art FP8 quantization to maximize performance. Notably, for Sailor-20B-Chat, we avoid quantizing the KV cache to FP8 due to significant performance degradation.}
        \label{fig:inference_latency_of_models}
    \end{minipage}    
\end{figure}

\textbf{Compass-v2 predominantly achieves lower inference latency than competing models across varying workloads.} To further contextualize Compass-v2's efficiency, we compare its inference latency against several open-source models of similar scale, including Qwen2.5-32B-Instruct, Sailor-20B-Chat and Qwen2.5-14B-Instruct, under identical experimental settings. Input sequences are sampled from the ShareGPT dataset and are either repeated or truncated to maintain a fixed length of 1,024 tokens. The maximum number of generated tokens is constrained to 256 to ensure consistency across evaluations. As summarized in Figure \ref{fig:inference_latency_of_models}, Compass-v2 achieves lower response times in most cases across varying workload intensities. This performance advantage underscores the effectiveness of our architectural optimizations, making Compass-v2 a more efficient solution for real-world deployment scenarios.

\subsection{Model Quantization}

To accelerate the inference of the Compass-v2 model, we initially adopt FP8 quantization techniques, similar to those used in Compass-v1. Specifically, FP8 quantization is applied to the majority of matrix multiplications within the model, including all linear layer parameters and most of the activations in the self-attention and MOE layers. To preserve precision, the LM head and MOE routing modules are excluded from quantization. However, our findings indicate that Compass-v2 is not computationally intensive for small- to medium-sized batch processing due to the fine-grained expert computations. This makes FP8 quantization suboptimal, as its impact on computational efficiency is limited and the de-quantization overhead introduces additional complexity. As the batch size increases, the number of activated experts grows rapidly, with memory access emerging as the primary bottleneck. To alleviate this constraint, we further reduce the model weight precision to 4-bit using the AWQ technique~\citep{lin2023awq} . Additionally, AWQ quantization offers the advantage of supporting a broader range of GPU architectures, whereas FP8 quantization is only available on Hopper architecture and later generations. We utilize an in-house inference framework to perform both offline calibration and online inference.

\textbf{Efficiency and Accuracy evaluation.} We evaluate the efficiency of FP8 and AWQ quantized inference in comparison to FP16. As presented in Table \ref{tab:compass-v2 quantization performance}\begin{table}[htbp]
    \centering    
     \caption{Comparison of inference speed for FP16, FP8, and AWQ-INT4 quantization. The evaluation is conducted using in-house datasets from chatbot service, with a concurrency of 256. }
    \begin{tabular}{l ll}
        \toprule
           System&Tokens/second  &Speed up \\
         \midrule
              Compass-v2-FP16&2817.82 &1.0x \\
 Compass-v2-FP8&3885.49 &1.38x\\
               Compass-v2-AWQ-INT4&4439.98 &1.58x\\
         \bottomrule
    \end{tabular}
    \label{tab:compass-v2 quantization performance}
\end{table}, AWQ quantization yields superior performance compared to FP16 and FP8 quantization. Additionally, We assess the impact of quantization on a range of standard benchmarks, including reasoning, question answering, reading comprehension, safety and math tasks. The results summarized in Table \ref{Table:Evaluations on standard benchmarks with inference API} demonstrate that both FP8 and AWQ quantization incur minimal precision loss compared to FP16.

\begin{table}[hbtp]

\centering
\caption{ Evaluation on standard benchmarks using the Compass-v2 API. FP8 and AWQ-INT4 quantization approach has negligible impact on the model’s quality.
}
\resizebox{\columnwidth}{!}{
\begin{tabular}{@{}cccccc@{}l}

\toprule
\textbf{}            & \textbf{Reasoning} & \textbf{Question Answer} & \textbf{Reading Comprehension} &  \textbf{Safety} &\textbf{Math}  \\ \midrule
Compass-v2-FP16&  0.7060&   0.8392& 0.8859  & 0.5149& 0.2932  \\ 
Compass-v2-FP8& 0.7055&   0.8304&   0.8838 & 0.5011&  0.2924  \\ 
Compass-v2-AWQ-INT4& 0.6987& 0.8312& 0.8823    & 0.5330& 0.2874    \\ 
\bottomrule
\end{tabular}
}

\label{Table:Evaluations on standard benchmarks with inference API}
\end{table}

\subsection{Future Plan}

In the future, we plan to further optimize the Compass-v2 model to alleviate the memory access bottleneck during inference. One key improvement will be the implementation of multi-head latent attention to replace the current grouped-query attention, which will help reduce the KV-cache size and enhance throughput. Additionally, we will explore overlapping the computation of shared experts and activated experts to improve efficiency. Furthermore, we aim to introduce structured outputs in the API, ensuring that model outputs reliably conform to developer-supplied JSON schemas.

\section{Evaluation}\label{chapter:evaluation}


In order to comprehensively evaluate Compass-v2 from the perspective of standard capabilities and practical applications, we used both open-source data and in-house data to score the model performance.
Our evaluation process involves comparisons with several well-known models, such as Qwen2.5, LLaMa3, GPT-4o, and DeepSeek-V3. All evaluations are conducted through automated methods, with manual reviews incorporated to address potentially problematic cases. For open-source benchmark datasets, we utilize the llm-evaluation-harness framework~\citep{eval-harness}. For in-house free-form benchmark dataset-- which often involve more open-ended or domain-specific tasks-- we apply GPT-4 as a evaluator to provide fair and high-quality judgments.

\subsection{Basic Setting}
To demonstrate the evaluation process, we list the hyperparameters used for each models.
\begin{itemize}[leftmargin=14pt, itemindent=0pt, itemsep=0pt, topsep=0pt]
  \item \textbf{Compass-v2}: max\_tokens=2048, seed=42, repetition\_penalty=1.05, top\_k=5, top\_p=0.85, temperature=0.4
  \item \textbf{Qwen2.5-14b}: repetition\_penalty=1.05, temperature=0.7, top\_p=0.8, top\_k=20
  \item \textbf{LLaMa3-8b}: max\_tokens=512, temperature=0.7, top\_p=0.9, top\_k=40, repetition\_penalty=1.00, seed=42, stream=false
  \item \textbf{Seallm-v3-7b}: repetition\_penalty=1.05, temperature=0.7
  \item \textbf{Sailor2 / Moonlight-16B-A3B-Instruct}: huggingface opensource generation config
  \item \textbf{GPT-4o / DeepSeek R1}: official API defaults
\end{itemize}

We also briefly list the datasets we used in the evaluation and their functions.
OpenBookQA~\citep{mihaylov2018can} is a dataset designed to mimic open-book exams, where answering requires both core science facts and broader commonsense knowledge. 
XCOPA~\citep{ponti2020xcopa} is a benchmark to evaluate the a model's ability to transfer commonsense reasoning across 11 languages from 11 families and several areas around the world.
MMLU~\citep{hendrycks2020measuring} is a benchmark designed to evaluate a model’s ability to recall and apply factual and conceptual knowledge acquired during pretraining.  
HellaSwag~\citep{zellers2019hellaswag} is a challenging commonsense reasoning dataset created using Adversarial Filtering (AF).
Shopping MMLU~\citep{jin2024shopping} is a multi-task benchmark for evaluating LLMs' e-commerce capabilities across shopping concept understanding, knowledge reasoning, user behavior alignment, and multilingual proficiency.
ECInstruct~\citep{peng2024ecellm} is a large-scale e-commerce benchmark with 116,528 carefully curated samples across 10 tasks.
GSM8K~\citep{cobbe2021gsm8k} is a dataset of school math word problems designed to evaluate multi-step reasoning ability. 
GPQA \citep{rein2024gpqa} is a graduate-Level Q\&A Benchmark, which is a challenging dataset designed to evaluate the capabilities of LLMs.
BoolQ~\citep{clark2019boolq} is a dataset with naturally occurring yes/no questions collected from real Google search queries. Each question is paired with a relevant Wikipedia passage and human-annotated answer. 
ARC~\citep{clark2018think} is a dataset containing multiple-choice science questions designed to benchmark advanced reasoning abilities.
RACE \citep{lai-etal-2017-race} dataset is a machine reading comprehension dataset consisting of 27,933 passages and 97,867 questions from English exams.
Belebele~\citep{bandarkar2023belebele} is a multilingual multiple-choice reading comprehension dataset designed to evaluate model performance across high-, medium-, and low-resource languages.
XStoryCloze~\citep{lin2021few} is a multilingual story cloze test dataset that evaluates models' narrative understanding by selecting the more plausible story ending from two options.
Winogender~\citep{zhao2018gender} is a benchmark dataset for evaluating gender bias in language models. 

\subsection{Southeast Multilingual Datasets Evaluation}

As shown in Table \ref{tab:Open-Source Multilingual Evaluation}, \textbf{Compass-v2 demonstrates robust multilingual capabilities across Indonesian, Malay, Portuguese, Thai, Vietnamese, and Traditional Chinese benchmarks, consistently delivering competitive performance while maintaining remarkable consistency}. 
The model particularly excels in commonsense reasoning tasks (e.g., XCOPA), achieving standout scores-- 0.856 (ID), 0.864 (VI), and 0.908 (ZH)-- that frequently surpass comparable models. Its knowledge retention ability (shown on MMLU) is also strong across languages, ranging from 0.853 on ID to 0.760 on TH. The model's strong performance is further reflected on HellaSwag, where the model demonstrates reliable commonsense understanding, and in OpenBookQA, where it shows strong and consistent factual knowledge retrieval. Notably, Compass-v2 maintains exceptional stability in performance across all evaluated languages, in contrast to larger models that often exhibit significant performance variance depending on the language. This suggests that Compass-v2 has gone through thoughtful optimization for multilingual deployment scenarios.

\begin{table}[h]
\centering
\caption{Southeast Multilingual Evaluation}
\footnotesize
\setlength{\tabcolsep}{3pt}
\renewcommand{\arraystretch}{0.85}
\resizebox{\textwidth}{!}{
\begin{tabular}{@{}cl|>{\columncolor{gray!10!white}}c>{\columncolor{blue!10!white}}ccccccccc@{}}
\toprule
\textbf{language} & \textbf{Benchmark} & \makecell{\textbf{Compass} \\ -\textbf{v1}} & \makecell{\textbf{Compass} \\ -\textbf{v2}} & \makecell{LLaMa3.1 \\ -8b} & \makecell{Sailor2 \\ -14b} & \makecell{Sailor2 \\ -20b} & \makecell{Qwen2.5 \\ -7b} & \makecell{Qwen2.5 \\ -14b} & \makecell{Qwen2.5 \\ -32b} & \makecell{Seallm \\ -7b} & \makecell{Moonlight \\ -16B-A3B} \\
\midrule
\multirow{3}{*}{ }&Architecture & Dense & MoE & Dense & Dense & Dense & Dense & Dense & Dense & Dense& MoE \\
&\# Activated Params & 13B & 5B & 8B & 14B & 20B & 7B & 14B & 32B & 7B & 3B \\
&\# Total Params & 13B & 30B & 8B & 14B & 20B & 7B & 14B & 32B & 7B & 16B \\
\midrule
\multirow{5}{*}{ID}&openbookqa & \textbf{0.424} & 0.418 & 0.364 & 0.404 & \underline{0.420} & 0.382 & 0.398 & \underline{0.420} & 0.362 & 0.328 \\
&xcopa & 0.782 & \textbf{0.856} & 0.710 & 0.766 & 0.782 & 0.732 & 0.786 & \underline{0.790} & 0.710 & 0.630 \\
&mmlu & 0.640 & \textbf{0.853} & 0.567 & 0.593 & 0.705 & 0.640 & 0.709 & \underline{0.760} & 0.632 & 0.507 \\
&hellaswag & 0.635 & \textbf{0.685} & 0.530 & 0.645 & \underline{0.680} & 0.570 & 0.620 & 0.620 & 0.545 & 0.480 \\
\midrule
\multirow{5}{*}{MY}&openbookqa & \textbf{0.412} & 0.380 & 0.320 & 0.358 & \underline{0.406} & 0.324 & 0.360 & 0.354 & 0.292 & 0.290 \\
&xcopa & 0.746 & \textbf{0.818} & 0.700 & 0.716 & \underline{0.748} & 0.638 & 0.662 & 0.688 & 0.668 & 0.598 \\
&mmlu & 0.643 & \textbf{0.846} & 0.540 & 0.588 & 0.677 & 0.630 & 0.702 & \underline{0.744} & 0.591 & 0.498 \\
&hellaswag & \underline{0.695} & \textbf{0.705} & 0.540 & 0.660 & 0.685 & 0.530 & 0.585 & 0.640 & 0.555 & 0.450 \\
\midrule
\multirow{5}{*}{PT}&openbookqa & \underline{0.452} & 0.436 & 0.436 & 0.402 & 0.446 & 0.444 & \underline{0.452} & \textbf{0.458} & 0.414 & 0.346 \\
&xcopa & 0.768 & \textbf{0.844} & 0.706 & 0.660 & 0.748 & 0.736 & 0.774 & \underline{0.780} & 0.716 & 0.662 \\
&mmlu & 0.668 & \textbf{0.856} & 0.600 & 0.605 & 0.719 & 0.674 & 0.777 & \underline{0.814} & 0.661 & 0.579 \\
&hellaswag & 0.690 & \textbf{0.745} & 0.615 & 0.535 & 0.670 & 0.625 & 0.660 & \underline{0.710} & 0.600 & 0.530 \\
\midrule
\multirow{5}{*}{TH}&openbookqa & 0.364 & 0.352 & 0.314 & 0.338 & \textbf{0.376} & 0.328 & 0.358 & \underline{0.366} & 0.336 & 0.284 \\
&xcopa & 0.624 & \textbf{0.718} & 0.588 & 0.646 & \underline{0.668} & 0.614 & 0.644 & 0.632 & 0.614 & 0.548 \\
&mmlu & 0.560 & \textbf{0.760} & 0.468 & 0.556 & 0.647 & 0.539 & 0.660 & \underline{0.683} & 0.523 & 0.360 \\
&hellaswag & \underline{0.650} & 0.600 & 0.530 & 0.610 & \textbf{0.670} & 0.520 & 0.590 & 0.625 & 0.540 & 0.370 \\
\midrule
\multirow{5}{*}{VI}&openbookqa & \textbf{0.358} & 0.330 & 0.300 & 0.320 & \underline{0.336} & 0.288 & 0.310 & 0.302 & 0.284 & 0.266 \\
&xcopa & 0.780 & \textbf{0.864} & 0.698 & 0.788 & 0.810 & 0.780 & \underline{0.828} & 0.812 & 0.724 & 0.602 \\
&mmlu & 0.612 & \textbf{0.796} & 0.516 & 0.602 & 0.698 & 0.621 & 0.711 & \underline{0.739} & 0.590 & 0.430 \\
&hellaswag & 0.645 & \textbf{0.705} & 0.545 & 0.625 & 0.670 & 0.575 & 0.660 & \underline{0.690} & 0.565 & 0.415 \\
\midrule
\multirow{5}{*}{\makecell{ZH\\(Traditional)}}&openbookqa & \textbf{0.414} & 0.394 & 0.352 & 0.344 & 0.402 & 0.366 & \underline{0.410} & 0.402 & 0.392 & 0.338 \\
&xcopa & 0.792 & \textbf{0.908} & 0.718 & 0.750 & 0.800 & 0.806 & \underline{0.830} & 0.824 & 0.766 & 0.762 \\
&mmlu & 0.591 & \underline{0.800} & 0.556 & 0.596 & 0.707 & 0.656 & 0.746 & \textbf{0.804} & 0.651 & 0.607 \\
&hellaswag & 0.630 & 0.660 & 0.620 & 0.630 & \underline{0.720} & 0.670 & \textbf{0.755} & \textbf{0.755} & 0.645 & 0.680 \\
\bottomrule
\end{tabular}
}
\label{tab:Open-Source Multilingual Evaluation}
\end{table}

\subsection{Ecommerce Datasets Evaluation}
Considering the application scenarios of Compass-v2, we carefully evaluated the model's capabilities on multiple open-source e-commerce datasets (i.e., Shopping MMLU and ECInstruct) to verify the practical applicability of the model.
From Table \ref{tab:Open-Source Ecommerce Field Evaluation}, we can conclude that \textbf{Compass-v2 demonstrates comprehensive strengths in e-commerce related tasks, achieving a best average score of 55.74\% that places it in the top tier.}

\begin{table}[h]
\centering
\caption{E-commerce Evaluation}
\footnotesize
\setlength{\tabcolsep}{3pt}
\renewcommand{\arraystretch}{0.85}
\resizebox{\textwidth}{!}{
\begin{tabular}{@{}l|>{\columncolor{gray!10!white}}c>{\columncolor{blue!10!white}}ccccccccc@{}}
\toprule
 \textbf{Benchmark} & \makecell{\textbf{Compass} \\ -\textbf{v1}} & \makecell{\textbf{Compass} \\ -\textbf{v2}} & \makecell{LLaMa3.1 \\ -8b} & \makecell{Sailor2 \\ -14b} & \makecell{Sailor2 \\ -20b} & \makecell{Qwen2.5 \\ -7b} & \makecell{Qwen2.5 \\ -14b} & \makecell{Qwen2.5 \\ -32b} & \makecell{Seallm \\ -7b} & \makecell{Moonlight \\ -16B-A3B} \\
\midrule

Architecture & Dense & MoE & Dense & Dense & Dense & Dense & Dense & Dense & Dense& MoE \\
\# Activated Params & 13B & 5B & 8B & 14B & 20B & 7B & 14B & 32B & 7B & 3B \\
\# Total Params & 13B & 30B & 8B & 14B & 20B & 7B & 14B & 32B & 7B & 16B \\
\midrule

 Ecom QA & 0.6943 & 0.8190 & 0.7883 & 0.7895 & \underline{0.8413} & 0.7842 & 0.8334 & \textbf{0.8486} & 0.8050 & 0.7940 \\
  Shopping Concept & \underline{0.5813} & \textbf{0.6205} & 0.4793 & 0.3202 & 0.4743 & 0.4821 & 0.5510 & 0.5624 & 0.3406 & 0.3938 \\
  Ecom Reasoning & \textbf{0.4218} & 0.3794 & 0.3517 & 0.2640 & 0.2513 & 0.2663 & \underline{0.4013} & 0.2823 & 0.2252 & 0.3095 \\
  User Understanding & \underline{0.3250} & 0.2817 & \textbf{0.3267} & 0.2684 & 0.2617 & 0.2267 & 0.1750 & 0.1283 & 0.1317 & 0.2617 \\
  Ecom Generation & \underline{0.6157} & \textbf{0.6865} & 0.4734 & 0.5512 & 0.5232 & 0.5899 & 0.5596 & 0.6026 & 0.5279 & 0.4910 \\
\midrule
  \textbf{Average} & \underline{0.5276} & \textbf{0.5574} & 0.4839 & 0.4387 & 0.4704 & 0.4698 & 0.5041 & 0.4849 & 0.4061 & 0.4499 \\
\bottomrule
\end{tabular}
}
\label{tab:Open-Source Ecommerce Field Evaluation}
\end{table}

The model shows outstanding performance in e-commerce Q\&A (81.90\%) and shopping concept understanding (62.05\%), significantly outperforming other open source models with similar size (below 10B). 
Compass-v2 achieves a 37.94\% score in challenging e-commerce reasoning tasks, while maintaining solid user-understanding capabilities (28.17\%). In addition, Compass-v2 also performs well in e-commerce content generation with a leading 68.65\% score - 7.08\% higher than its closest competitor. 
This balanced yet specialized performance highlights Compass-v2's advantages in various e-commerce scenarios, delivering superior content generation capabilities while maintaining parameter efficiency, making it an excellent choice for practical e-commerce AI applications.

\subsection{General Datasets Evaluation}

We carefully select several well-recognized open-source benchmarks to evaluate Compass-v2 across multiple aspects in English including reasoning, safety, mathematics, question answer, reading comprehension and coding.

\begin{table}[h]
\centering
\caption{General Ability Evaluation}
\footnotesize
\setlength{\tabcolsep}{3pt}
\renewcommand{\arraystretch}{0.85}
\resizebox{\textwidth}{!}{
\begin{tabular}{@{}l|>{\columncolor{gray!10!white}}c>{\columncolor{blue!10!white}}ccccccccc@{}}
\toprule
\textbf{Benchmark} & \makecell{\textbf{Compass} \\ -\textbf{v1}}& \makecell{\textbf{Compass} \\ -\textbf{v2}} & \makecell{LLaMa3.1 \\ -8b} & \makecell{Sailor2 \\ -14b} & \makecell{Sailor2 \\ -20b} & \makecell{Qwen2.5 \\ -7b} & \makecell{Qwen2.5 \\ -14b} & \makecell{Qwen2.5 \\ -32b} & \makecell{Seallm \\ -7b} & \makecell{Moonlight \\ -16B-A3B} \\
\midrule

Architecture&Dense & MoE & Dense & Dense & Dense & Dense & Dense & Dense & Dense & MoE \\
\# Activated Params &13B& 5B & 8B & 14B & 20B & 7B & 14B & 32B & 7B & 3B \\
\# Total Params &13B& 30B & 8B & 14B & 20B & 7B & 14B & 32B & 7B & 16B \\
\midrule

GSM8K &0.6262& \underline{0.8150} & 0.7680 & 0.6619 & 0.7597 & \textbf{0.8347} & 0.8006 & 0.7574 & 0.7938 & 0.7794 \\
MMLU &0.7159& 0.7801 & 0.6836 & 0.6634 & 0.7755 & 0.7427 & \underline{0.7991} & \textbf{0.8323} & 0.7017 & 0.6949 \\
GPQA &0.2778& \underline{0.4091} & 0.3283 & 0.2424 & 0.3485 & 0.3283 & \textbf{0.4394} & 0.3838 & 0.3182 & 0.3384 \\
BoolQ &0.8709& \underline{0.8826} & 0.8419 & 0.7728 & 0.8789 & 0.8636 & 0.8801 & \textbf{0.8954} & 0.8492 & 0.8471 \\
ARC&0.5734 & 0.5913 & 0.5520 & 0.5273 & \textbf{0.6425} & 0.5503 & \underline{0.6220} & 0.5896 & 0.4923 & 0.5853 \\
HellaSwag &0.8179& \textbf{0.8580} & 0.7927 & 0.7705 & 0.8251 & 0.8050 & 0.8433 & \underline{0.8517} & 0.7889 & 0.8044 \\
OpenBookQA &0.4660& \textbf{0.5000} & 0.4300 & 0.4200 & 0.4740 & \underline{0.4840} & 0.4780 & 0.4640 & 0.4400 & 0.4480 \\
Social &0.5067& \textbf{0.5425} & 0.4939 & 0.4719 & 0.5148 & \underline{0.5174} & \textbf{0.5425} & 0.5015 & 0.4780 & 0.5046 \\
RACE &\textbf{0.5263}& 0.4842 & 0.4459 & 0.3694 & 0.4478 & 0.4612 & 0.4708 & \underline{0.4852} & 0.3770 & 0.4134 \\
XCOPA &0.8680& \textbf{0.9400} & 0.8400 & 0.8360 & 0.8860 & 0.8480 & 0.8780 & \underline{0.8880} & 0.8540 & 0.8500 \\
Belebele &0.9000& \textbf{0.9667} & 0.8867 & 0.8244 & 0.9289 & 0.9111 & 0.9411 & \underline{0.9500} & 0.8822 & 0.8789 \\
XStoryCloze &0.8233& \textbf{0.8438} & 0.7968 & 0.7750 & 0.8213 & 0.8041 & \underline{0.8259} & 0.8094 & 0.7895 & 0.7975 \\
Winogender &0.6722& 0.7736 & 0.6389 & 0.7347 & 0.8139 & 0.6875 & \underline{0.8208} & \textbf{0.8333} & 0.6167 & 0.6736 \\
\midrule
\textbf{Average} &0.6650& \textbf{0.7221} & 0.6537 & 0.6207 & 0.7013 & 0.6798 & \underline{0.7186} & 0.7109 & 0.6447 & 0.6627\\
\bottomrule
\end{tabular}
}
\label{tab:Open-Source General Field Evaluation}
\end{table}

\textbf{Compass-v2 has demonstrated strong capabilities in the general English field, surpassing most models of the same level.}
From the evaluation results in Table \ref{tab:Open-Source General Field Evaluation}, 
we can see that the model consistently ranks in the top three across all benchmarks, demonstrating strong general-performance compared to all evaluated baselines.
In mathematical reasoning, Compass-v2 scores 81.5\% on GSM8K, outperforming all peers except Qwen2.5-7b. Compass-v2 also maintains strong language understanding with a BoolQ score of 88.26\%, closely rivaling larger models like Qwen2.5-32b (89.54\%). Its performance in commonsense reasoning is particularly noteworthy, scoring 85.8\% on HellaSwag, which represents a significant 6.5-point lead over similar-sized models like LLaMa3.1-8b (79.27\%).
The model also shows impressive results  in multi-task knowledge evaluation, wtih an MMLU score of 78.01\%, surpassing most competitors in its parameter size.
For more challenging tasks, such as graduate-level reasoning on GPQA, Compass-v2 scores 40.91\%, maintaining a clear lead over  comparable models (LLaMa3.1-8b: 32.83\%). In scientific reasoning (ARC: 59.13\%) and social interaction understanding (Social: 54.25\%), Compass-v2 demonstrates balanced and robust performance, narrowing the gap with larger model in these domains. 

Notably, Compass-v2 achieves this comprehensive capability profile while maintaining the best average score (72.21\%) among all evaluated models, also improved by 8.59\% compared to Compass-v1, highlighting its exceptional computational efficiency.

\subsection{In-house Datasets Evaluation}
The currently available open-source evaluation datasets mainly consist of multiple-choice and true/false questions, the free-form, open-ended queries commonly posed by real users in practical scenarios. To address this, we have constructed a series of in-house evaluation datasets to comprehensively assess the performance of LLMs in multilingual capabilities.
Each evaluation question is provided in multiple language versions, including English, Indonesian, Malay, Portuguese, Thai, Vietnamese, and Traditional Chinese. We conduct a comprehensive evaluation of the model from a SEA multilingual perspective.




\begin{table}[h]
\centering
\caption{In-House Datasets Evaluation}
\footnotesize
\setlength{\tabcolsep}{3pt}
\renewcommand{\arraystretch}{0.85}
\resizebox{\textwidth}{!}{
\begin{tabular}{@{}c|>{\columncolor{blue!10!white}}ccccccc@{}}

\textbf{Benchmark} & \textbf{Compass-v2} & \textbf{Qwen2.5-14b} & \textbf{Qwen2.5-32b} & \textbf{Qwen2.5-72b} & \textbf{LLaMa3-70b} & \textbf{GPT-4o} & \textbf{DeepSeek V3} \\
\toprule

Architecture & MoE & Dense & Dense & Dense & Dense & - & MoE \\
\# Activated Params &  5B &  14B & 32B & 72B & 70B & - & 37B \\
\# Total Params & 30B& 14B & 32B & 72B & 70B & - & 671B \\
\midrule

English & 79.37 & 76.63 & 79.53 & 81.83 & 80.35 & 79.64 & \textbf{84.10} \\
Indonesian & 78.19 & 73.66 & 75.34 & 77.24 & 74.84 & 77.29 & \textbf{79.15} \\
Malay & 76.41 & 74.55 & 77.01 & 77.94 & 74.99 & 79.15 & \textbf{79.32} \\
Portuguese & \textbf{75.95} & 67.88 & 71.82 & 72.94 & 70.76 & 71.88 & 74.52 \\
Thai & 73.95 & 74.27 & \textbf{77.94} & 77.56 & 73.84 & 77.45 & 76.46 \\
\makecell{Chinese\\(Traditional)} & 76.74 & 75.97 & 77.89 & 79.04 & 77.61 & \textbf{79.20} & 78.55 \\
Vietnamese & 75.90 & 75.70 & \textbf{79.47} & 78.71 & 75.04 & 79.15 & 77.92 \\
\midrule
\textbf{Average} & 76.64 & 74.09 & 77.00 & 77.89 & 75.35 & 77.68 & \textbf{78.57} \\
\bottomrule
\end{tabular}
}
\label{Table:In-House_Multilingual_Results}
\end{table}

\textbf{Despite being smaller in scale, Compass-v2 achieves comparable performance to the industry's top models on the in-house dataset.}
From the evaluation results shown in Table \ref{Table:In-House_Multilingual_Results}, Compass-v2 achieves a balanced performance with an average score of 76.64 across all evaluated languages. Compared to significantly larger models like Qwen2.5-72b (77.89 avg) and GPT-4o (77.68 avg), Compass-v2 delivers reliable multilingual performance at a more efficient scale.
The model shows remarkable consistency, excelling in Indonesian (78.19) and maintaining competitive scores in English (79.37), Malay (76.41), and Chinese (76.74). 
Its robust performance in SEA languages like Thai (73.95) and Vietnamese (75.90) further reflects its ability to effectively handle linguistically diverse and low-resource inputs.
Notably, Compass-v2 beats all the top models in Portuguese with a score of 75.95.
Compass-v2's balanced performance-- combined with its efficiency-- makes it well-suited for real-world applications requiring stable performance across multiple language environments, especially in SEA scenarios.

\section{Related Work}\label{sec:related-work}

\subsection{General Large Language Models}
Recent advancements in natural language processing (NLP) have been driven largely by the rapid development of Large Language Models (LLMs), which leverage transformer architectures and massive-scale pre-training on diverse text corpora. Among the most influential are the GPT series, developed by OpenAI, which demonstrated the impressive capabilities in language understanding and generation tasks. Following GPT's pionneering success, several open-source models have emerged, such as the LLaMA~\citep{Dubey2024TheL3}, Qwen~\citep{Yang2024Qwen25TR}, DeepSeek~\citep{DeepSeekAI2024DeepSeekV3TR}, OLMo~\citep{OLMo20242O2}, MAP-Neo~\citep{Zhang2024MAPNeoHC}.
These models are typically pre-trained on massive and diverse corpora and serve as foundational models for a wide variety of downstream applications through fine-tuning or instruction tuning. However, most existing LLMs are heavily focused on high-resource languages such as English and Chinese, and often lack adequate support for low-resource languages, including many spoken in Southeast Asia.

\subsection{SEA Language Models}
There has been some recent progress in creating Southeast Asian (SEA) language models. A common approach is to continue pre-training existing foundation models on SEA corpora. This method leverages the general capabilities of high-resource LLMs while adapting them to SEA languages and domains. For example, the SEA-LION series~\citep{sea_lion_2024} and SeaLLM series~\citep{Nguyen2023SeaLLMsL} have taken the initiatives to undergo continues pretraining to create SEA language models. However, these models still fail to achieve a performance level comparable to that of commercial models, such as GPT-4o~\citep{openai2024gpt4technicalreport}. To better boost models performance on SEA languages. the Sailor series~\citep{sailor2report} further improve open models to support 13 SEA languages. Although these continued pre-training models can rapidly adapt to regional contexts, they may still inherit the limitations of the original model architecture or tokenizer, which may not be optimal for morphologically rich Southeast Asian languages.

Training LLMs from scratch specifically for Southeast Asian languages is sill relatively rare due to high cost of data collection and computation resources. However, some recent efforts have emerged to address this gap, e.g. SEA-LION-v1~\citep{sea_lion_2024}. However, these models for Southeast Asia remain limited in scale and scope compared to general-purpose LLMs.

\section{Conclusion}

In this paper, we introduced Compass-v2, a fine-grained Mixture-of-Experts (MoE) model with shared experts. By carefully crafting multilingual data, leveraging MoE and a train-from-scratch approach, Compass-v2 effectively captures linguistic diversity and commercial nuances in Southeast Asian languages while maintaining high efficiency.

Our multi-stage pre-training and post-training strategy, including long-context extension and direct preference optimization (DPO), allows the model to achieve strong multilingual, e-commerce, and business-related performance, outperforming similarly sized models. Through quantization techniques such as AWQ and FP8, we further optimize inference efficiency, enabling practical deployment in real-world applications.

Comprehensive evaluations show that Compass-v2 delivers competitive performance with significantly fewer active parameters, achieving high efficiency-to-performance trade-offs. Our deployment through Shopee’s CAP platform ensures seamless access via API and web-based interfaces, facilitating integration across various business workflows.

Moving forward, we plan to expand Compass-v2’s multilingual coverage, enhance its reasoning capabilities, and further optimize scalability for enterprise adoption. Our work highlights the potential of sparse-activated architectures in building efficient, scalable, and domain-adaptive large language models, paving the way for future innovations in multilingual and industry-specific AI applications.
\section{Disclaimer}

This paper is published as a public service for general informational purposes only.  Shopee Limited and its affiliates (“Shopee”) is not responsible for the content or accuracy of any information contained in this paper, and shall not be responsible for any decisions made by another person based on such information.  Shopee makes no representations or warranties that the data or information presented in this paper is correct or sufficient to support the conclusions reached or that the research design or methodology used to reach such conclusions is adequate.

\clearpage

\bibliography{main}
\bibliographystyle{plainnat}
\clearpage


\end{document}